\documentclass[10pt,twocolumn,letterpaper]{article}

\usepackage{cvpr}
\usepackage{times}
\usepackage{epsfig}
\usepackage{graphicx}
\usepackage{subcaption}
\usepackage{colortbl}
\usepackage{amsmath}
\usepackage{amssymb}
\usepackage{booktabs}
\usepackage{pifont}
\usepackage{multirow}
\usepackage{dblfloatfix}
\usepackage[dvipsnames]{xcolor}
\usepackage[title]{appendix}
\usepackage{comment}
\usepackage{enumitem}
\definecolor{dcolor}{HTML}{5b00a7}

\newcommand{\emoji}{\raisebox{-1pt}{\includegraphics[height=10px]{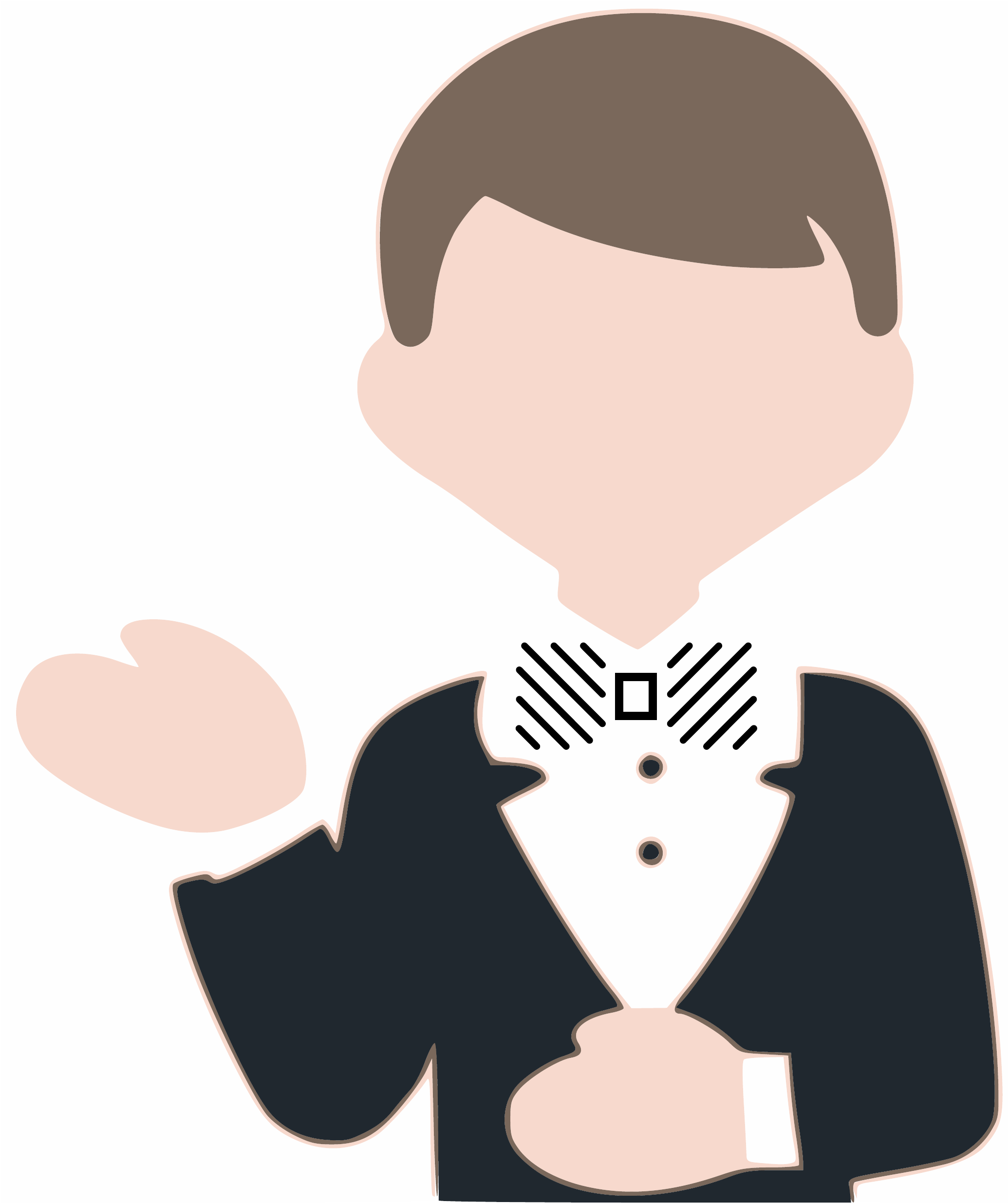}}}
\newcommand{\emojibig}{\raisebox{-1pt}{\includegraphics[height=15px]{imgs/butler_bowtie.pdf}}}

\newcommand{\dataset}{\textsc{ALFRED}}
\newcommand{\datasetlong}{\textbf{A}ction \textbf{L}earning \textbf{F}rom \textbf{R}ealistic \textbf{E}nvironments and \textbf{D}irectives}
\newcommand{\datasetemoji}{\textbf{\dataset{}}}

\newcommand{\overbar}[1]{\mkern 1.5mu\overline{\mkern-1.5mu#1\mkern-1.5mu}\mkern 1.5mu}

\newcommand{\emphasis}[1]{\emph{#1}}
\newcommand{\objects}[1]{\textit{#1}}
\newcommand{\langquote}[1]{``#1''}
\newcommand{\verbs}[1]{\texttt{#1}}
\newcommand{\scenes}[1]{\textsc{#1}}
\newcommand{\splits}[1]{\textit{#1}}
\newcommand{\tasks}[1]{\textbf{#1}}
\newcommand{\basemodels}[1]{\textsc{#1}}

\newcommand{\cmark}{\raisebox{0pt}{\color{blue}{\ding{51}}}}
\newcommand{\xmark}{\raisebox{0pt}{\color{red}{\ding{55}}}}
\newcommand{\good}[1]{\textcolor{blue}{\textbf{#1}}}
\newcommand{\bad}[1]{\textcolor{red}{\textbf{#1}}}
\definecolor{purplish}{RGB}{99, 57, 116}
\newcommand{\neutral}[1]{\textcolor{purplish}{\textbf{#1}}}

\renewcommand{\eqref}[1]{(Equation~\ref{#1})}
\newcommand{\figref}[1]{Figure~\ref{#1}}

\newcommand{\tabref}[1]{Table~\ref{#1}}

\newcommand{\thor}{AI2\nobreakdash-THOR}

\usepackage[hang,flushmargin]{footmisc}

\usepackage[pagebackref=true,breaklinks=true,letterpaper=true,colorlinks,bookmarks=false]{hyperref}

\cvprfinalcopy 

\title{\vspace{-0pt}\dataset{} \emojibig{}\\ A Benchmark for Interpreting Grounded Instructions for Everyday Tasks}

\author{\vspace{-30pt}\\
 Mohit Shridhar$^1$\hspace{3em} Jesse Thomason$^1$\hspace{3em} Daniel Gordon$^1$\hspace{3em} Yonatan Bisk$^{1,2,3}$\\
 Winson Han$^3$\hspace{3.1em} Roozbeh Mottaghi$^{1,3}$\hspace{3.1em} Luke Zettlemoyer$^1$\hspace{3.1em} Dieter Fox$^{1,4}$\\
  \textbf{\url{AskForALFRED.com}} \\
}

\date{}

\begin{document}


\maketitle

\begin{abstract}
  We present \textbf{\dataset{}} ({\normalfont\datasetlong{}}), a benchmark for learning a mapping from natural language instructions and egocentric vision to sequences of actions for household tasks.
  \dataset{} includes long, compositional tasks with non-reversible state changes to shrink the gap between research benchmarks and real-world applications.
  \dataset{} consists of expert demonstrations in interactive visual environments for 25k natural language directives.
  These directives contain both high-level goals like \langquote{Rinse off a mug and place it in the coffee maker.} and low-level language instructions like \langquote{Walk to the coffee maker on the right.}
  \dataset{} tasks are more complex in terms of sequence length, action space, and language than existing vision-and-language task datasets.
  We show that a baseline model based on recent embodied vision-and-language tasks performs poorly on \dataset{}, suggesting that there is significant room for developing innovative grounded visual language understanding models with this benchmark.
\end{abstract}

\footnotetext[1]{
Paul G. Allen School of Computer Sci. \& Eng., Univ. of Washington\\[-15pt]

\hspace{-3pt}$^2$Language Technologies Institute @ Carnegie Mellon University\\[-15pt]

\hspace{-3pt}$^3$Allen Institute for AI\hspace{1em}$^4$NVIDIA
}

\vspace{-2em}
\section{Introduction}

\begin{figure}[t]
    \centering
    \includegraphics[width=\linewidth]{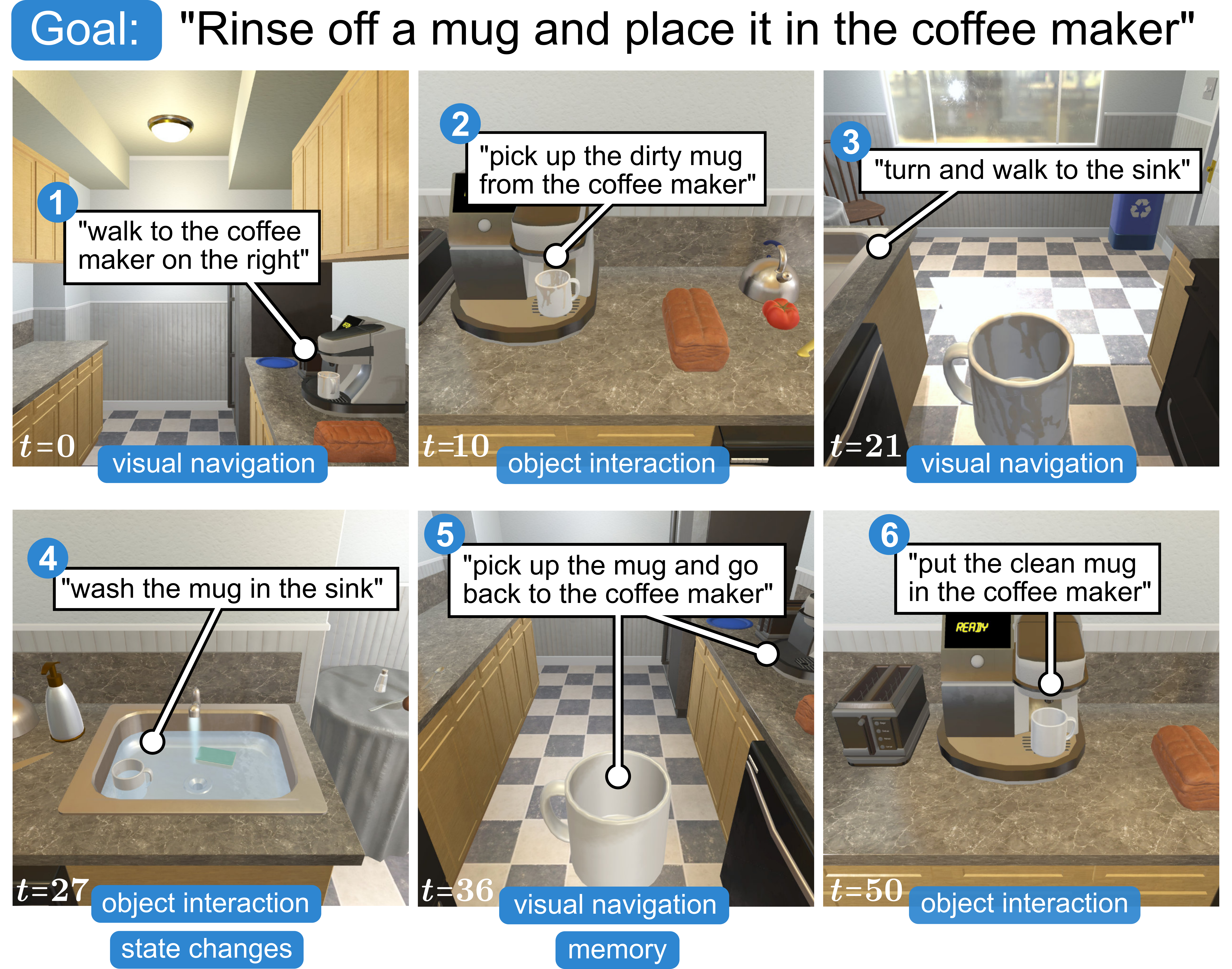}
    \caption{
    \dataset{} consists of 25k language directives corresponding to expert demonstrations of household tasks.
    We highlight several frames corresponding to portions of the accompanying language instruction.
    \dataset{} involves interactions with objects, keeping track of state changes, and references to previous instructions.}
    \label{fig:teaser}
    \vspace{-1em}
\end{figure}

A robot operating in human spaces must learn to connect natural language to the world.
This \emphasis{symbol grounding}~\cite{harnad:phys90} problem has largely focused on connecting language to static images.
However, robots need to understand task-oriented language, for example \langquote{Rinse off a mug and place it in the coffee maker,} as illustrated in Figure~\ref{fig:teaser}.

Platforms for translating language to action have become increasingly popular, spawning new test-beds~\cite{anderson18,chaplot2017,chen2019,VirtualHome}.
These benchmarks include language-driven navigation and embodied question answering, which have seen dramatic improvements in modeling thanks to environments like Matterport 3D~\cite{anderson18,Matterport3D}, \thor{}~\cite{ai2thor}, and AI Habitat~\cite{Habitat}. However, these datasets ignore complexities arising from describing task-oriented behaviors with objects.

We introduce \datasetemoji{}, a new benchmark for connecting human language to \emphasis{actions}, \emphasis{behaviors}, and \emphasis{objects} in interactive visual environments.
Planner-based expert demonstrations are accompanied by both high- and low-level human language instructions in 120 indoor scenes in \thor~2.0~\cite{ai2thor}.
These demonstrations involve partial observability, long action horizons, underspecified natural language, and irreversible actions.

\begin{table*}
\centering
\begin{small}
\begin{tabular}{@{}lcccccccc@{}}
& \multicolumn{2}{c}{\bf --- Language ---} & \multicolumn{3}{c}{\bf --- Virtual Environment ---} & \multicolumn{3}{c}{\bf --- Inference ---} \\
& \# Human&  \multirow{2}{*}{Granularity} & Visual & Movable & State   & \multirow{2}{*}{Vis. Obs.} &  \multirow{2}{*}{Navigation} &  \multirow{2}{*}{Interaction} \\
&  Annotations &             & Quality & Objects & Changes &     &            &             \\
\toprule
TACoS \cite{tacos:regnerietal:tacl} & \good{17k+} & \good{High\&Low} & \good{Photos} & \xmark & \xmark & --  & -- & -- \\
R2R \cite{anderson18}; Touchdown \cite{chen2019} & \good{21k+}\textbf{;} \neutral{9.3k+} & \neutral{Low} & \good{Photos} & \xmark & \xmark & \good{Ego} & \neutral{Graph} & \xmark \\
EQA \cite{eqa} & \xmark & \neutral{High} & \bad{Low} & \xmark & \xmark & \good{Ego} & \good{Discrete} & \xmark \\
Matterport EQA \cite{wijmans2019embodied} & \xmark & \neutral{High} & \good{Photos} & \xmark & \xmark & \good{Ego} & \good{Discrete} & \xmark \\
IQA \cite{iqa} & \xmark & \neutral{High} & \good{High} & \xmark & \cmark & \good{Ego} & \good{Discrete} & \neutral{Discrete} \\
VirtualHome \cite{VirtualHome} & \neutral{2.7k+} & \good{High\&Low} & \good{High} & \cmark & \cmark & \bad{3\textsuperscript{rd} Person} & \xmark & \neutral{Discrete} \\
VSP \cite{zhu2017} & \xmark & \neutral{High} & \good{High} & \cmark & \cmark & \good{Ego} & \xmark & \neutral{Discrete} \\
\midrule          
\multirow{2}{*}{\dataset{} \emoji{}} & \multirow{2}{*}{\good{25k+}} & \multirow{2}{*}{\good{High\&Low}} & \multirow{2}{*}{\good{High}} & \multirow{2}{*}{\cmark} & \multirow{2}{*}{\cmark} & \multirow{2}{*}{\good{Ego}} & \multirow{2}{*}{\good{Discrete}} & \good{Discrete} \\
                               &        &                  &            &      & & & &   \good{+ Mask} \\
\bottomrule
\end{tabular}
\label{fig:dataset_comparison}
\end{small}
\caption{\textbf{Dataset comparison.} \dataset{} is the first interactive visual dataset to include high-level goal and low-level natural language instructions for object and environment interactions.
TACoS~\cite{tacos:regnerietal:tacl} provides detailed high- and low-level text descriptions of cooking videos, but does not facilitate task execution.
For navigation, \dataset{} enables discretized, grid-based movement, while other datasets use topological graph navigation or avoid navigation altogether.
\dataset{} requires an agent to generate spatially located interaction masks for action commands.
By contrast, other datasets only require choosing from a discrete set of available interactions and object classes or offer no interactive capability.
}
\label{tab:datasets}
\end{table*}

\dataset{} includes 25,743 English language directives describing 8,055 expert demonstrations averaging 50 steps each, resulting in 428,322 image-action pairs.
Motivated by work in robotics on segmentation-based grasping~\cite{mousavian2019graspnet}, agents in \dataset{} interact with objects visually, specifying a pixelwise interaction mask of the target object.
This inference is more realistic than simple object class prediction, where localization is treated as a solved problem.
Existing beam-search~\cite{fried2018,tan2019,wang2019} and backtracking solutions~\cite{ke2019,ma2019b} are infeasible due to the larger action and state spaces, long horizon, and inability to undo certain actions.

To establish baseline performance levels, we evaluate a sequence-to-sequence model akin to existing vision-and-language navigation tasks~\cite{ma2019a}.
This model is not effective on the complex tasks in \dataset{}, achieving less than 5\% success rates.
For analysis, we also evaluate individual sub-goals.
While performance is better for isolated sub-goals, the model lacks the reasoning capacity for long-horizon and compositional task planning.

In summary, \dataset{} facilitates learning models that translate from language to sequences of actions and interactions in a visually and physically realistic simulation environment.
This benchmark captures many challenges present in real-world settings for translating human language to robot actions for accomplishing household tasks.
Models that can overcome these challenges will begin to close the gap towards real-world, language-driven robotics.

\section{Related Work}

Table~\ref{tab:datasets} summarizes the benefits of \dataset{} relative to other visual action datasets with language annotations.

\begin{figure*}[!ht]
\begin{center}
\includegraphics[width=\linewidth]{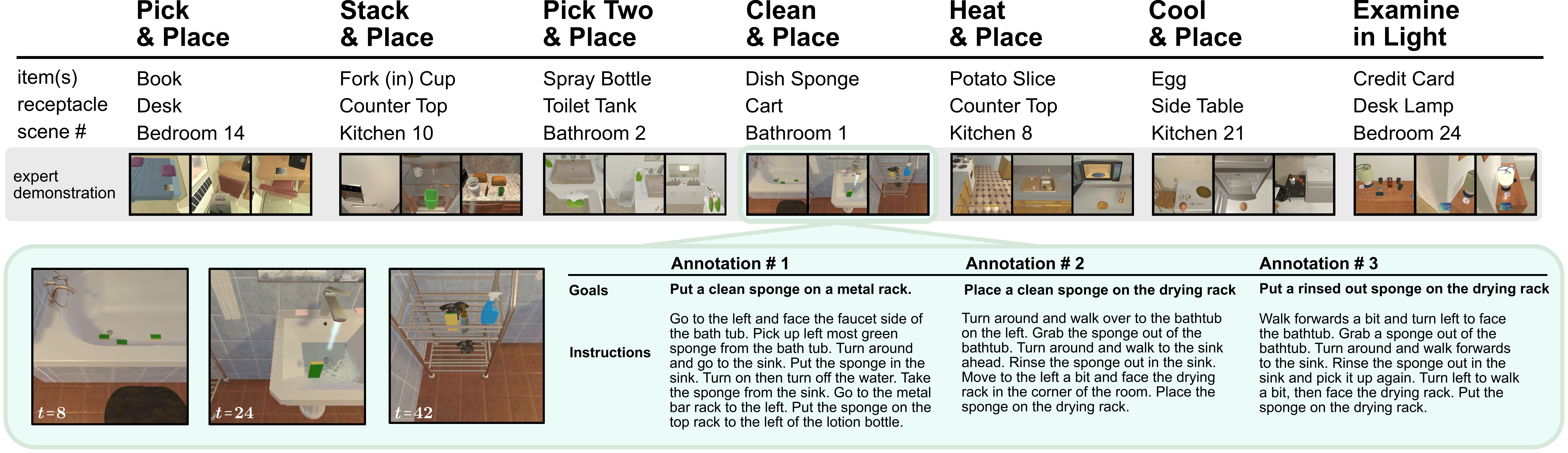}
\caption{\textbf{\dataset{} annotations.}
We introduce 7 different task types parameterized by 84 object classes in 120 scenes.
An example of each task type is given above.
For the \textbf{Clean \& Place} demonstration, we also show the three crowdsourced language directives. Please see the supplemental material for example demonstrations and language for each task.}
\label{fig:task_examples}
\end{center} 
\end{figure*}

\noindent \textbf{Vision \& Language Navigation.} In vision-and-language navigation tasks, either natural or templated language describes a route to a goal location through egocentric visual observations~\cite{anderson18,chaplot2017,chen:aaai11,chen2019,macmahon:aaai06}.
Since the proposal of R2R~\cite{anderson18}, researchers have dramatically improved the navigation performance of models~\cite{fried2018,ke2019,ma2019b,wang2019,wang2018} with techniques like progress monitoring~\cite{ma2019a}, as well as introduced task variants with additional, on-route instructions~\cite{nguyen:emnlp19,nguyen2019,thomason:corl19}.
Much of this research is limited to static environments.
By contrast, \dataset{} tasks include navigation, object interactions, and state changes.  

\noindent \textbf{Vision \& Language Task Completion.} There are several existing benchmarks based on simple block worlds and fully observable scenes~\cite{Bisk:2016ab,misra2017}.
\dataset{} provides more difficult tasks in richer, visually complex scenes, and uses partially observable environments.
The CHAI benchmark~\cite{misra2018} evaluates agents performing household instructions, but uses a generic ``interact'' action.
\dataset{} has seven manipulation actions, such as pick up, turn on, and open, state changes like clean versus dirty, and variation in language and visual complexity.

Previous work in the original \thor{} environment investigated the task of visual semantic planning~\cite{gordon2018hiprl,zhu2017}.
Artificial language came from templates, and environment interaction was handled with discrete class predictions, for example selecting \objects{apple} as the target object from predefined options.
\dataset{} features human language instructions, and object selections are carried out with class-agnostic, pixelwise interaction masks.
In VirtualHome~\cite{VirtualHome}, programs are generated from video demonstration and natural language instructions, but inference does not involve egocentric visual and action feedback or partial observability.

There is an extensive literature on language-based instruction following in the natural language processing community.
There, research has focused on mapping instructions to actions~\cite{artzi2013,chen:aaai11,malmaud2014,misra2015,tenorth2010}, but these works do not involve visual, interactive environments.

\noindent \textbf{Embodied Question Answering.} Existing datasets for visual question answering in embodied environments use templated language or static scenes~\cite{eqa,iqa,wijmans2019embodied,eqa_multitarget}.
In \dataset{}, rather than answering a question, the agent must complete a task specified using natural language, which requires both navigation and interaction with objects.


\noindent \textbf{Instruction Alignment.} Language annotations of videos enable discovering visual correspondences between words and concepts~\cite{alayrac2016,sener2015,tacos:regnerietal:tacl,yu2013,zhukov2019}.
\dataset{} requires performing tasks in an interactive setting as opposed to learning from recorded videos.

\noindent \textbf{Robotics Instruction Following.} Instruction following is a long-standing topic of interest in robotics~\cite{pancakes2011,bollini2012,macglashan2015,misra2014,nyga2018,paul2018,shridhar2018interactive,thomason2015}.
Lines of research consider different tasks such as cooking~\cite{bollini2012}, table clearing~\cite{nyga2018}, and mobile manipulation~\cite{macglashan2015}.
In general, they are limited to a few scenes~\cite{misra2014}, consider a small number of objects~\cite{macglashan2015}, or use the same environment for training and testing~\cite{pancakes2011}.
In contrast, \dataset{} includes 120 scenes, many object classes with diverse appearances, 
and a test set of unseen environments.

\section{The \dataset{} Dataset}
\label{sec:dataset}
The \dataset{} dataset comprises 25,743 language directives corresponding to 8,055 expert demonstration episodes.
Each directive includes a high-level goal and a set of step-by-step instructions.
Each expert demonstration can be deterministically replayed in the \thor{} 2.0 simulator.

\subsection{Expert Demonstrations}
Expert demonstrations are composed of an agent's egocentric visual observations of the environment and what action is taken at each timestep as well as ground-truth interaction masks. These demonstrations are generated by a planner~\cite{hoffmann2001ff} using metadata not available to the agent at inference time. Navigation actions move the agent or change its camera orientation, while manipulation actions include picking and placing objects, opening and closing cabinets and drawers, and turning appliances on and off.
Interactions can involve multiple objects, such as using a knife to slice an apple, cleaning a mug in the sink, and heating a potato in the microwave.
Manipulation actions are accompanied by a ground truth segmentation of the target object.

Figure~\ref{fig:task_examples} gives examples of the high-level agent tasks in \dataset{}, like putting a cleaned object at a destination.
These tasks are parameterized by the object of focus, the destination receptacle (\eg, \objects{table top}), the scene in which to carry out the task, and in the case of \tasks{Stack \& Place}, a base object (\eg, \objects{plate}).
\dataset{} contains expert demonstrations of these seven tasks executed using combinations of 58 unique object classes and 26 receptacle object classes across 120 different indoor scenes.
For object classes like \objects{potato slice}, the agent must first pick up a \objects{knife} and find a \objects{potato} to create slices.
All object classes contain multiple visual variations with different shapes, textures, and colors.
For example, there are 30 unique variants of the \objects{apple} class.
Indoor scenes include different room types: 30 each of kitchens, bathrooms, bedrooms, and living rooms.

For 2,685 combinations of task parameters, we generate three expert demonstrations per parameter set, for a total of 8,055 unique demonstrations with an average of 50 action steps.
The distributions of actions steps in \dataset{} demonstrations versus related datasets is given in Figure~\ref{fig:distributions}.
As an example, for task parameters \{task: \tasks{Heat \& Place}, object: \objects{potato}, destination: \objects{counter top}, scene: \scenes{kitchen-8}\}, we generate three different expert demonstrations by starting the agent and objects in randomly chosen locations.
Object start positions have some commonsense, class-specific constraints, for example a \objects{fork} can start inside a \objects{drawer}, but an \objects{apple} cannot.

Contrasting navigation-only datasets where expert demonstrations can come from an $A^*$ planner, our state space includes object positions and state changes.
Thus, to generate expert demonstrations we encode the agent and object states, as well as high-level environment dynamics, into Planning Domain Definition Language (PDDL) rules~\cite{pddl}.
We then define task-specific PDDL goal conditions, for example that a heated \objects{potato} is resting on a \objects{table top}.
Note that the planner encodes the environment as \emphasis{fully observable} and has perfect knowledge about world dynamics.
For training and testing agent models, however, the environment is \emphasis{partially observable}: it is only viewed through the agent's egocentric vision as actions are carried out.

\begin{table}
\centering
\begin{small}
\begin{tabular}{@{}lrrrrr@{}}
 & \bf Train & \multicolumn{2}{c}{\bf Validation} & \multicolumn{2}{c}{\bf Test}    \\
& & \splits{Seen} & \splits{Unseen} & \splits{Seen} & \splits{Unseen} \\
\toprule
\# Annotations & 21,023 & 820 & 821 & 1,533 & 1,529 \\
\# Scenes & 108 & 88 & 4 & 107 & 8 \\
\bottomrule
\end{tabular}
\end{small}
\caption{\textbf{\dataset{} Data Splits.} All expert demonstrations and associated language directives in the validation and test folds are distinct from those in the train fold.
The validation and test sets are split into \splits{seen} and \splits{unseen} folds.
Scenes in the \splits{seen} folds of validation and test data are subsets of those in the train fold.
Scenes in the \splits{unseen} validation and test folds are distinct from the train folds and from each other.}
\label{tab:data_splits}
\end{table}

We split these expert demonstrations into training, validation, and test folds (Table~\ref{tab:data_splits}).
Following work in vision-and-language navigation~\cite{anderson18}, we further split the validation and test into two conditions: \splits{seen} and \splits{unseen} environments.
This split facilitates examining how well models generalize to entirely new spaces with novel object class variations.

\subsection{Language Directives} \label{sseq:language_directives}

\begin{figure}
\includegraphics[width=\linewidth]{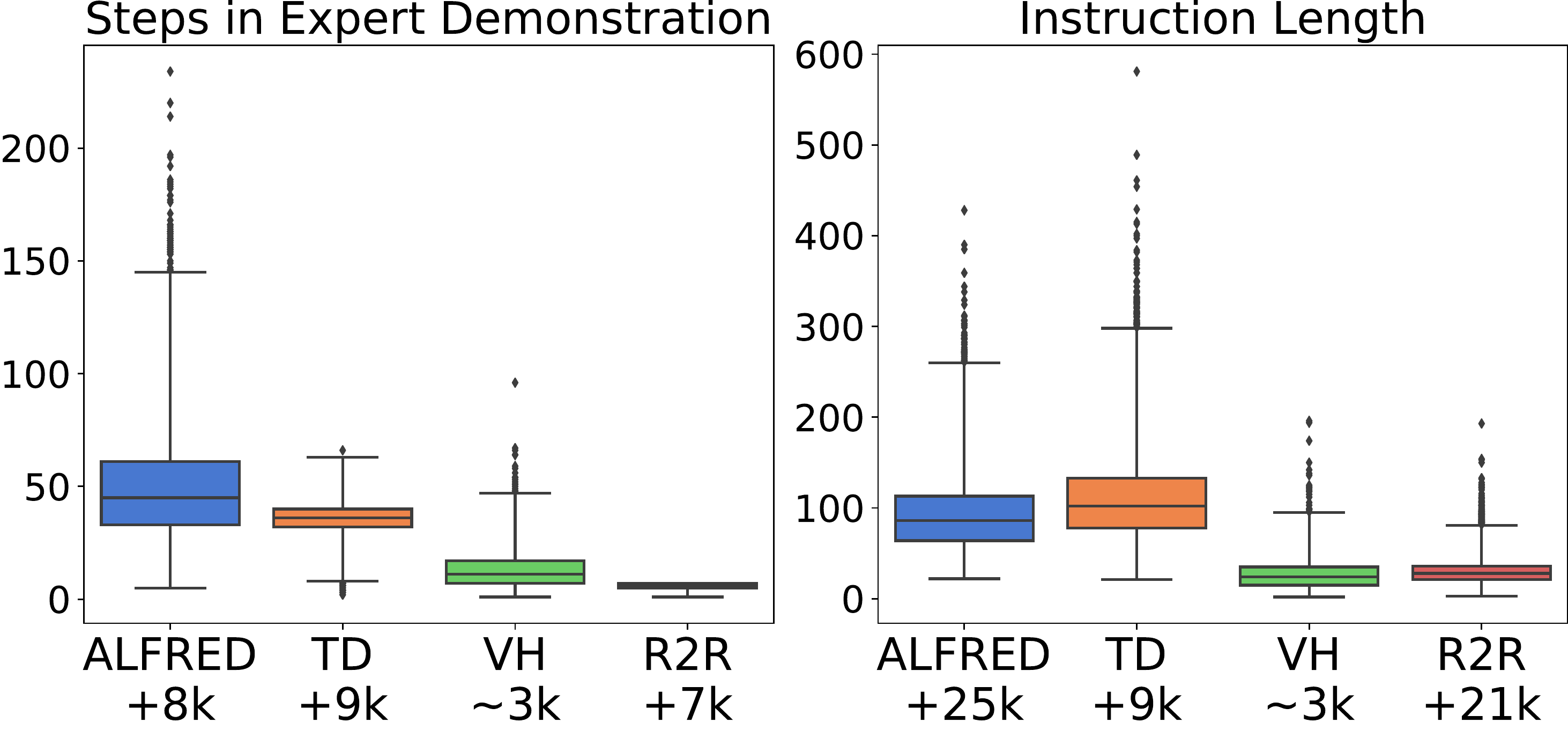}
\caption{\textbf{Comparison to Existing Datasets.} Expert demonstration steps and instruction tokens of \dataset{} compared to other datasets with human language for action sequences: Touchdown (TD)~\cite{chen2019}, VirtualHome (VH)~\cite{VirtualHome}, and Room-to-Room (R2R)~\cite{anderson18}.
The total number of demonstrations or annotations is given with the dataset label.}
\label{fig:distributions}
\end{figure}

For every expert demonstration, we collect open vocabulary, free-form language directives from at least three different annotators using Amazon Mechanical Turk (AMT), resulting in 25k total language directives.
Language directives include a high-level \emphasis{goal} together with low-level \emphasis{instructions}, as shown in Figures~\ref{fig:teaser} and \ref{fig:task_examples}.
The distribution of language annotation token lengths in \dataset{} versus related datasets is given in Figure~\ref{fig:distributions}.

AMT workers are told to write instructions to tell a ``smart robot'' how to accomplish what is shown in a video.
We create a video of each expert demonstration and segment it such that each segment corresponds to an instruction.
We consult the PDDL plan for the expert demonstration to identify task sub-goals, for example the many low-level steps to navigate to a knife, or the several steps to heat a potato slice in the microwave once standing in front of it.
We visually highlight action sequences related to sub-goals via colored timeline bars below the video.
In each HIT (Human Intelligence Task), a worker watches the video, then writes low-level, step-by-step \emphasis{instructions} for each highlighted sub-goal segment.
The worker also writes a high-level \emphasis{goal} that summarizes what the robot should accomplish during the expert demonstration.

These directives are validated through a second HIT by at least two annotators, with a possible third tie-breaker.
For validation, we show a worker all three language directive annotations without the video.
The worker selects whether the three directives describe the same actions, and if not, which is most different.
If a directive is chosen as most different by a majority of validation workers, it is removed and the demonstration is subsequently re-annotated by another worker.
Qualitatively, these rejected annotations contain incorrect object referents (\eg, \langquote{egg} instead of \langquote{potato}) or directions (\eg, \langquote{go left towards...} instead of \langquote{right}).

\begin{figure*}[t]
    \centering
    \includegraphics[width=\linewidth]{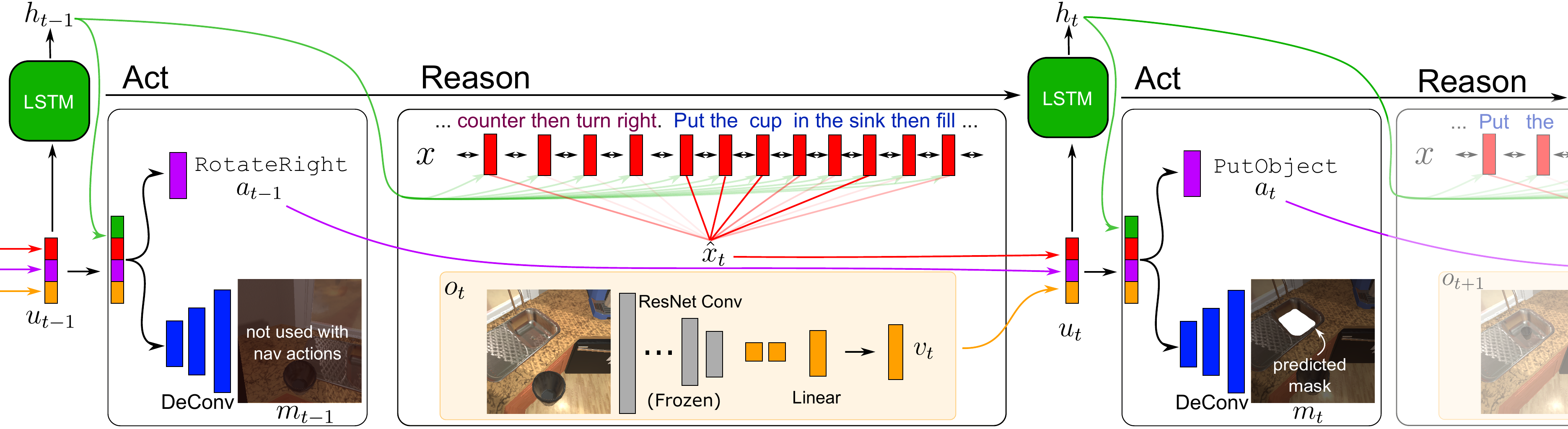}
    \caption{\textbf{Model overview.} At each step, our model reweights the instruction based on the history ($\hat{x}_t$), and combines the current observation features ($v_t$) and the previously executed action ($a_{t-1}$).  These are passed as input to an LSTM cell to produce the current hidden state.  Finally, the new hidden state ($h_t$) is combined with the previous features to predict both the next action ($a_t$) and a pixelwise interaction mask over the observed image to indicate an object.
    }
    \label{fig:model}
\end{figure*}

\section{Baseline Models}

An agent trained for \dataset{} tasks needs to jointly reason over vision and language input and produce a sequence of low-level actions to interact with the environment.

\subsection{Sequence-to-Sequence Models}

We model the interactive agent with a CNN-LSTM sequence-to-sequence (\basemodels{Seq2Seq}) architecture.
A CNN enodes the visual input, a bidirectional-LSTM generates a representation of the language input, and a decoder LSTM infers a sequence of low-level actions while attending over the encoded language.
See~\figref{fig:model} for an overview and the supplementary material for implementation details.

\vspace{-3pt}
\paragraph{Supervision.}
We train all models using imitation learning on expert trajectories. This ensures the language directives match the visual inputs.
At each timestep, the model is trained to produce the expert action and associated interaction mask for manipulation actions.

We note that a DAgger-style~\cite{ross2011reduction} student-forcing paradigm in \dataset{} is non-trivial, even disregarding language alignment.
Obtaining expert demonstration actions on the fly in navigation-only datasets like R2R~\cite{anderson18} only requires rerunning $A^*$.
In \dataset{}, on the fly demonstrations requires re-planning.
In same cases re-planning is not possible: if during a task of \{\tasks{Clean \& Place}, \objects{apple}, \objects{refrigerator}, \scenes{kitchen-3}\} a student-forcing model slices the only \objects{apple} in the scene, the action cannot be recovered from and the task cannot be completed.

\vspace{-3pt}
\paragraph{Visual encoding.}
Each visual observation $o_{t}$ is encoded with a frozen ResNet-18~\cite{He_2016_CVPR} CNN, where we take the output of the final convolution layer to preserve spatial information necessary for grounding specific objects in the visual frame. We embed this output using two more $1\times1$ convolution layers and a fully-connected layer.
During training, a set of $T$ observations from the expert demonstration is encoded as $\overbar{V} = \langle v_{1}, v_{2}, \dots, v_{T} \rangle$, where $v_{t}$ is the visual feature vector at time-step $t$.

\vspace{-3pt}
\paragraph{Language encoding.}
Given a natural language goal $\overbar{G} = \langle g_{1}, g_{2}, \dots g_{L_{g}} \rangle$ of $L_{g}$ words, and step-by-step instructions $\overbar{S} = \langle s_{1}, s_{2} \dots s_{L_{s}} \rangle$ of $L_{s}$ words, we append them into a single input sequence $\overbar{X} = \langle g_{1}, g_{2}, \dots g_{L_{g}}, \small{\texttt{<SEP>}}, s_{1}, s_{2} \dots s_{L_{s}} \rangle$ with the $\small{\texttt{<SEP>}}$ token indicating the separation between the high-level goal and low-level instructions.
This sequence is fed into a bi-directional LSTM encoder to produce an encoding $x = \{x_{1}, x_{2}, \dots, x_{L_{g} + L_{s}} \}$ for each word in $\overbar{X}$.

\vspace{-3pt}
\paragraph{Attention over language.}
The agent's action at each timestep is based on an attention mechanism weighting tokens in the instruction.
We perform soft-attention on the language features $x$ to compute the attention distribution $\alpha_{t}$ conditioned on the hidden state of the decoder $h_{t-1}$ from the last timestep:
\begin{equation}
\begin{split}
    z_{t} &= (W_{x} h_{t-1})^\top x, \\[4pt] 
    \alpha_{t} &= \textrm{Softmax}(z_{t}), \\[4pt]
    \hat{x}_{t} &= \alpha_{t}^\top x
\end{split}
\end{equation}

\noindent
where $W_{x}$ are learnable parameters of a fully-connected layer, $z_{t}$ is a vector of scalar values that represent the attention mass for each word in $x$, and $\hat{x}_{t}$ is the weighted sum of $x$ over the attention distribution $\alpha_{t}$ induced from $z_{t}$.

\vspace{-3pt}
\paragraph{Action decoding.} At each timestep $t$, upon receiving a new observation image $o_{t}$, the LSTM decoder takes in the visual feature $v_{t}$, language feature $\hat{x}_{t}$, and the previous action $a_{t-1}$, and outputs a new hidden state $h_{t}$:
\begin{equation}
\begin{split}
    u_{t} &= [v_{t}; \hat{x}_{t}; a_{t-1}], \\[4pt]
    h_{t} &= \textrm{LSTM} \ (u_{t}, h_{t-1}) 
\end{split}
\end{equation}

\noindent
where $[;]$ denotes concatenation.
The hidden state $h_{t}$ is used to obtain the attention weighted language feature $\hat{x}_{t+1}$.

\vspace{-3pt}
\paragraph{Action and mask prediction.}
The agent interacts with the environment by choosing an action and producing a pixelwise binary mask indicating a specific object in the frame.
Although \thor{} supports continuous control for agent navigation and object manipulation, we discretize the action space.
The agent chooses from among 13 actions.
There are 5 navigation actions: $\texttt{MoveAhead}$, $\texttt{RotateRight}$, $\texttt{RotateLeft}$, $\texttt{LookUp}$, and $\texttt{LookDown}$ together with 7 interaction actions: $\texttt{Pickup}$, $\texttt{Put}$, $\texttt{Open}$, $\texttt{Close}$, $\texttt{ToggleOn}$, $\texttt{ToggleOff}$, and $\texttt{Slice}$.
Interaction actions require a pixelwise mask to denote the object of interest.\footnote{The final object chosen by the interaction API is based on the Intersection-over-Union (IoU) score between the predicted mask and the ground-truth object mask from the simulator.}
Finally, the agent predicts a $\texttt{Stop}$ action to end the episode.
We concatenate the hidden state $h_{t}$ with the input features $u_{t}$ and train two separate networks to predict the next action $a_{t}$ and interaction mask $m_{t}$:
\begin{equation}
\begin{split}
    a_{t} &= \textrm{argmax} \ (W_{a} \ [ h_{t}; u_{t}] ) \ , \\[4pt]
    m_{t} &= \sigma \ ( \small{\textrm{\textbf{deconv}}} \ [ h_{t}; u_{t} ] )
\end{split}
\end{equation}

\noindent
where $W_{a}$ are learnable parameters of a fully connected layer, \textbf{deconv} is a three-layer deconvolution network, and $\sigma$ is a sigmoid activation function.
Action selection is trained using softmax cross entropy with the expert action. The interaction masks are learned end-to-end in a supervised manner based on ground-truth object segmentations using binary cross-entropy loss.
The mask loss is rebalanced to account for sparsity in these dense masks in which target objects can take up a small portion of the visual frame.

\subsection{Progress Monitors}

\dataset{} tasks require reasoning over long sequences of images and instruction words.
We propose two auxiliary losses (Eq. \ref{eq:progress_monitor} \& \ref{eq:subgoal_progress}) that use additional temporal information to reduce this burden and form a sequence-to-sequence model with progress monitoring (\basemodels{Seq2Seq+PM}).




Ma \etal~\cite{ma2019a} showed that agents benefit from maintaining an internal estimate of their progress towards the goal for navigation tasks.
Akin to learning a value function in reinforcement learning, progress monitoring helps to learn the utility of each state in the process of achieving the overall task.
Intuitively, this allows our agent to better distinguish between visually similar states such as just before putting an object in the microwave versus just after taking the object out.
We introduce a simple module that predicts progress, $p_{t} \in [0, 1]$, conditioned on the decoder hidden state $h_{t}$ and the concatenated input $u_{t}$:
\begin{equation}
    p_{t} = \sigma \ ( W_{p} \ [h_{t}; u_{t}] ).
    \label{eq:progress_monitor}
\end{equation}
The supervision for $p_{t}$ is based on normalized time-step values $t/T$, where $t$ is the current time-step, and $T$ is the total length of the expert demonstration (trained via L2 loss).

We also train the agent to predict the number of sub-goals completed so far, $c_{t}$.
These sub-goals represent segments in the demonstration corresponding to sequences of actions like navigation, pickup, and heating as identified in the PDDL plan, discussed in Section~\ref{sseq:language_directives}.
Each segment has a corresponding language instruction, but the alignment must be learned.
This sub-goal prediction encourages the agent to coarsely track its progress through the language directive. 
This prediction is also conditioned on the decoder hidden state $h_{t}$ and the concatenated input $u_{t}$:
\begin{equation}
    c_{t} = \sigma \ ( W_{c} \ [h_{t}; u_{t}] ).
    \label{eq:subgoal_progress}
\end{equation}
We train $c_{t}$ in a supervised fashion by using the normalized number of sub-goals accomplished in the expert trajectory at each timestep, $c_t / C$, as the ground-truth label for a task with $C$ sub-goals.
We again train with an L2 loss.

\newcommand{\mcc}[2]{\multicolumn{#1}{c}{#2}}
\newcommand{\mcp}[2]{\multicolumn{#1}{c@{\hspace{30pt}}}{#2}}
\definecolor{Gray}{gray}{0.90}
\newcolumntype{a}{>{\columncolor{Gray}}r}
\newcolumntype{b}{>{\columncolor{Gray}}c}

\newcommand{\B}[1]{\textcolor{blue}{\textbf{#1}}}

\begin{table*}
    \setlength{\aboverulesep}{0pt}
    \setlength{\belowrulesep}{0pt}
    \centering
    \begin{small}
    \begin{tabular}{@{}laarrcaarr@{}}
                     & \mcp{4}{\textbf{Validation}} & \hspace{0pt} & \mcc{4}{\textbf{Test}}                       \\
                     & \mcc{2}{\splits{Seen}}   & \mcc{2}{\splits{Unseen}}  
                     & 
                     & \mcc{2}{\splits{Seen}}   & \mcc{2}{\splits{Unseen}}  \\
    Model            & \multicolumn{1}{b}{Task} & \multicolumn{1}{b}{Goal-Cond} 
                     & \multicolumn{1}{c}{Task} & \multicolumn{1}{c}{Goal-Cond} 
                     & 
                     & \multicolumn{1}{b}{Task} & \multicolumn{1}{b}{Goal-Cond} 
                     & \multicolumn{1}{c}{Task} & \multicolumn{1}{c}{Goal-Cond} \\
    \addlinespace[-\aboverulesep] 
    
    \cmidrule[\heavyrulewidth]{1-5}
    \cmidrule[\heavyrulewidth]{7-10}
    \basemodels{No Language}        & 0.0 (0.0)    & 5.9  (3.4)    & 0.0 (0.0)   & 6.5 (4.7)   & & 0.2 (0.0)   & 5.0 (3.2)   & 0.2 (0.0) & 6.6 (4.0) \\[1pt]
    \basemodels{No Vision}      & 0.0 (0.0)    & 5.7  (4.7)    & 0.0 (0.0)   & 6.8 (6.0)   & & 0.0 (0.0)   & 3.9 (3.2)   & 0.2 (0.1) & 6.6 (\B{4.6}) \\[1pt]
    \basemodels{Goal-only}          & 0.1 (0.0)    & 6.5  (4.3)    & 0.0 (0.0)   & 6.8 (5.0)   & & 0.1 (0.1)   & 5.0 (3.7)   & 0.2 (0.0) & 6.9 (4.4) \\[1pt]
    \basemodels{Instructions-only}  & 2.3 (1.1)    & 9.4  (6.1)    & 0.0 (0.0)   & \B{7.0} (4.9)   & & 2.7 (1.4)   & 8.2 (5.5)   & \B{0.5} (\B{0.2}) & 7.2 (\B{4.6}) \\[1pt]
    \cmidrule{1-5}  
    \cmidrule{7-10}  
    \basemodels{Seq2Seq}                    & 2.4 (1.1)    &  9.4 (5.7)    & \B{0.1} (0.0)   & 6.8 (4.7)      & & 2.1 (1.0)   & 7.4 (4.7)    & \B{0.5} (\B{0.2}) & 7.1 (4.5) \\[1pt]
\hspace{2pt}+ \basemodels{PM Progress-only} & 2.1 (1.1)    &  8.7 (5.6)    & 0.0 (0.0)   & 6.9 (5.0)      & & 3.0 (1.7)   & 8.0 (5.5)    & 0.3 (0.1) & \B{7.3} (4.5) \\[1pt]
\hspace{2pt}+ \basemodels{PM SubGoal-only}  & 2.1 (1.2)    &  9.6 (5.5)    & 0.0 (0.0)   & 6.6 (4.6)      & & 3.8 (1.7)   & 8.9 (5.6)    & \B{0.5} (\B{0.2}) & 7.1 (4.5)  \\[1pt]
    \hspace{2pt}+ PM Both         & \B{3.7} (\B{2.1})     &  \B{10.0} (\B{7.0})    & 0.0 (0.0)   & 6.9 (\B{5.1})  & & \B{4.0} (\B{2.0})   & \B{9.4} (\B{6.3})    & 0.4 (0.1) & 7.0 (4.3) \\[1pt]
    \cmidrule{1-5} \cmidrule{7-10}
    \basemodels{Human}     & \multicolumn{1}{b}{-} & \multicolumn{1}{b}{-} & \multicolumn{1}{c}{-} & \multicolumn{1}{c}{-} & & \multicolumn{1}{b}{-} & \multicolumn{1}{b}{-} & 91.0 (85.8) & 94.5 (87.6) \\
    \addlinespace[-\belowrulesep] 
    \cmidrule[\heavyrulewidth]{1-5}
    \cmidrule[\heavyrulewidth]{7-10}
    \end{tabular}
    \end{small}
    \caption{\textbf{Task and Goal-Condition Success.}
    For each metric, the corresponding path weighted metrics are given in parentheses.
    The highest values per fold and metric are shown in \B{blue}. All values are percentages.}
    \vspace{-10pt}
    \label{tab:results}
\end{table*}

\section{Experiments} 
\label{sec:experiements}

We evaluate the baseline models in the \thor{} simulator.
When evaluating on test folds, we run models with the lowest validation loss.
Episodes that exceed $1000$ steps or cause more than $10$ failed actions are terminated.
Failed actions arise from bumping into walls or predicting action interaction masks for incompatible objects, such as attempting to \texttt{Pickup} a \textit{counter top}.
These limitations encourage efficiency and reliability.
We assess the overall and partial success of models' task executions across episodes.

\subsection{Evaluation Metrics}
\dataset{} allows us to evaluate both full task and task goal-condition completion.
In navigation-only tasks, one can only measure how far the agent is from the goal.
In \dataset{}, we can also evaluate whether task goal-conditions have been completed, for example that a \objects{potato} has been sliced.
For all of our experiments, we report both Task Success and Goal-Condition Success.
Each Goal-Condition relies on multiple instructions, for example navigating to an object and then slicing it.
 
\paragraph{Task Success.}
Each expert demonstration is parameterized by a task to be performed, as illustrated in \figref{fig:task_examples}.
Task Success is defined as $1$ if the object positions and state changes correspond correctly to the task goal-conditions at the end of the action sequence, and $0$ otherwise.
Consider the task: \langquote{Put a hot potato slice on the counter}.
The agent succeeds if, at the end of the episode, any \objects{potato slice} object has changed to the \objects{heated} state and is resting on any \objects{counter top} surface.

\paragraph{Goal-Condition Success.}
The goal-condition success of a model is the ratio of goal-conditions completed at the end of an episode to those necessary to have finished a task.
For example, in the previous \tasks{Heat \& Place} example, there are four goal-conditions.
First, a \objects{potato} must be sliced.
Second, a \objects{potato slice} should become \objects{heated}.
Third, a \objects{potato slice} should come to rest on a \objects{counter top}.
Fourth, the same \objects{potato slice} that is \objects{heated} should be on the \objects{counter top}.
If the agent slices a \objects{potato}, then moves a slice to the counter top without heating it, then the goal-condition success score is $2/4=50\%$.
On average, tasks in \dataset{} have $2.55$ goal conditions.
The final score is calculated as the average goal-condition success of each episode.
Task success is $1$ only if goal-condition success is $1$.

\paragraph{Path Weighted Metrics.}
We include a Path Weighted version of both metrics that considers the length of the expert demonstration~\cite{spl}.
Expert demonstrations found via a PDDL solver on global information are not guaranteed to be optimal.
However, they avoid exploration, use shortest path navigation, and are generally efficient.
The path weighted score $p_s$ for metric $s$ is given as 
\vspace{-5pt}
\begin{equation}
    p_s = s \times \frac{L^*}{max(L^*, \hat{L})}
\vspace{-5pt}
\end{equation}
where $\hat{L}$ is the number of actions the model took in the episode, and $L^*$ is the number of actions in the expert demonstration.
Intuitively, a model receives half-credit for taking twice as long as the expert to accomplish a task.

\subsection{Sub-Goal Evaluation}

Completing the entire sequence of actions required to finish a task is challenging.
In addition to assessing full task success, we study the ability of a model to accomplish the next sub-goal conditioned on the preceding expert sequence.
The agent is tested by first forcing it to follow the expert demonstration to maintain a history of states leading up to the sub-goal, then requiring it to complete the sub-goal conditioned on the entire language directive and current visual observation.
For the task \langquote{Put a hot potato slice on the counter} for example, we can evaluate the sub-goal of navigating to the potato after using the expert demonstration to navigate to and pick up a \objects{knife}.
The tasks in \dataset{} contain on average $7.5$ such sub-goals (results in Table \ref{tab:subtask_abblations}).

\section{Analysis}

Results from our experiments are presented in \tabref{tab:results}.
We find that the initial model, without spatial or semantic maps, object segmentations, or explicit object-state tracking, performs poorly on \dataset{}'s long-horizon tasks with high-dimensional state-spaces.
The \basemodels{Seq2Seq} model achieves $\sim$8\% goal-condition success rate, showing that the agent does learn to partially complete some tasks.
This headroom (as compared with humans) motivates further research into models that can perform the complex vision-and-language planning introduced by \dataset{}.
The performance starkly contrasts other vision-and-language datasets focused on navigation, where sequence-to-sequence with progress monitoring performs well~\cite{ma2019a}.

\subsection{Random Agent}
A random agent is commonly employed as a baseline in vision-and-language tasks.
In \dataset{}, an agent that chooses a uniform random action and generates a uniform random interaction mask at each timestep achieves 0\% on all folds, even without an API failure limit.

\subsection{Unimodal Ablations}
Previous work established that learned agents without visual inputs, language inputs, or both performed better than random agents and were competitive with initial baselines for several navigation and question answering tasks~\cite{Thomason:2019}.
These performance gaps were due to structural biases in the datasets or issues with model capacity.
We evaluate these ablation baselines (\basemodels{No Language} and \basemodels{No Vision}) to study vision and language  bias in \dataset{}.

The unimodal ablation performances in Table~\ref{tab:results} indicate that both vision and language  modalities are necessary to accomplish the tasks in \dataset{}. 
The \basemodels{No Language} model finishes some goal-conditions by interacting with familiar objects seen during training.
The \basemodels{No Vision} model similarly finishes some goal-conditions by following low-level language instructions for navigation and memorizing interaction masks for common objects like \objects{microwaves} that are centered in the visual frame. 

\subsection{Model Ablations}

We additionally ablate the amount of language supervision available to the model, as language directives are given as both a high-level goal and step-by-step instructions.
Providing only high-level, underspecified goal language (\basemodels{Goal-only}) is insufficient to complete the tasks, but is enough to complete some goal-conditions.
Using just low-level, step-by-step instructions (\basemodels{Instructions-only} performs similarly to using both high- and low-levels.
Thus, this simple model does not seem to exploit the goal instruction to plan out sub-goals for step-by-step execution.

The two progress monitoring signals are marginally helpful, increasing the success rate from $\sim$1\% to $\sim$2\%.
Progress monitoring leads to more efficient task completion, as indicated by the consistently higher path weighted scores.
They may help avoid action repetition and with the prediction of the \verbs{Stop} action.

The agent takes more steps than the expert in all cases, as indicated by the lower path weighted scores.
Sometimes, this is caused by failing to keep track of state-changes, for example heating up an \objects{egg} in the \objects{microwave} multiple times.
Further, the models also do not generalize well to \splits{unseen} scenes, due to the overall visual complexity in \dataset{} arising from new scenes and novel object class instances.

\subsection{Human evaluation}

We obtained a human evaluation of 100 randomly sampled directives from the \splits{unseen} test fold. 
The experiment involved 5 participants who completed 20 tasks each using a keyboard-and-mouse interface. 
Before the experiment, the participants were allowed to familiarize themselves with \thor{}. 
The action-space and task restrictions were identical to that of the baseline models. Overall, the participants obtained a high success rate of 91\%, while taking slightly longer than the expert with 86\% path-length weighted success rate. This indicates that the directives in \dataset{} are well-aligned with the demonstrations.   


\subsection{Sub-Goal Performance} \label{sec:subgoal_results}
We also examine performance of the \basemodels{Seq2Seq} model on individual sub-goals in \dataset{}. For this experiment, we use the expert trajectory to move the agent through the episode up to the sub-task.
Then, the agent begins inference based on the language directive and current visual frame.

\begin{table}[]
\begin{center}
\begin{small}
\begin{tabular}{l@{\hspace{4pt}}l@{\hspace{10pt}}r@{\hspace{7pt}}r@{\hspace{7pt}}r@{\hspace{7pt}}r@{\hspace{7pt}}r@{\hspace{7pt}}r@{\hspace{7pt}}r@{\hspace{7pt}}r@{\hspace{7pt}}r@{\hspace{7pt}}r}
&& \multicolumn{8}{@{}c@{}}{Sub-Goal Ablations - \textbf{Validation}} \\
&Model & \rotatebox{75}{Goto} & \rotatebox{75}{Pickup} & \rotatebox{75}{Put}  & \rotatebox{75}{Cool} & \rotatebox{75}{Heat} & \rotatebox{75}{Clean} & \rotatebox{75}{Slice} & \rotatebox{75}{Toggle} & Avg. \\ 
\toprule
\multirow{3}{*}{\rotatebox{90}{\splits{Seen}}}&No Lang         & 28     & 22      & 71     & \B{89}   & \B{87}  & 64      & 19      & 90      & 59    \\
&S2S            & 49     & 32      & 80     & 87       & 85      & \B{82}  & 23      & 97      & 67    \\
&S2S + PM   & \B{51} & \B{32}  & \B{81} & 88       & 85      & 81      & \B{25}  & \B{100} & \B{68}    \\ 
\midrule
\multirow{3}{*}{\rotatebox{90}{\splits{Unseen}}}&No Lang  & 17     & 9       & 31     & 75       & 86      & 13      & 8       & 4       & 30    \\
&S2S         & 21     & 20      & \B{51} & \B{94}   & 88      & 21      & \B{14}  & \B{54}  & 45    \\
&S2S + PM   & \B{22} & \B{21}  & 46     & 92       & \B{89}  & \B{57}  & 12      & 32      & \B{46} \\
\bottomrule
\end{tabular}
\end{small}
\end{center}
\caption{\textbf{Evaluations by path weighted sub-goal success.}
All values are percentages. The highest values per fold and task are shown in \B{blue}.
We note that the \basemodels{No Vision} model achieves less than 2\% on all sub-goals.
See supplemental material for more.}
\label{tab:subtask_abblations}
\end{table}

\tabref{tab:subtask_abblations} presents path-length weighted success scores for 8 sub-goals.
\tasks{Goto} and \tasks{Pickup} sub-tasks with the \basemodels{Seq2Seq+PM} model achieve $\sim$51\% and $\sim$32\%, respectively, even in \splits{seen} environments.
Visual semantic navigation is considerably harder in \splits{unseen} environments.
Similarly, interaction masks for \tasks{Pickup} actions in \splits{unseen} environments are worse due to unfamiliar scenes and object instances.
Simple sub-goals like \tasks{Cool}, and \tasks{Heat} are achieved at a high success rate of $\sim$90\% because these tasks are mostly object-agnostic.
For example, the agent becomes familiar with using microwaves to heat things regardless of the object in-hand, because microwaves have little visual diversity across kitchens. 
Overall, the sub-goal evaluations indicate that models that exploit modularity and hierarchy, or make use of pretrained object segmentation models, may make headway on full task sequences.

\section{Conclusions}

We introduced \dataset{}\emoji{}, a benchmark for learning to map natural language instructions and egocentric vision to sequences of actions. \dataset{} moves us closer to a community goal of language-driven robots  capable of navigation and interaction.
The environment dynamics and interaction mask predictions required in \dataset{} narrow the gap between what is required of agents in simulation and robots operating in the real world~\cite{mousavian2019graspnet}.

We use \dataset{} to evaluate a sequence-to-sequence model with progress monitoring, shown to be effective in other vision-and-language navigation tasks~\cite{ma2019a}.
While this model is relatively competent at accomplishing some sub-goals (\eg operating microwaves is similar across \textbf{Heat \& Place} tasks), the overall task success rates are poor.
The long horizon of \dataset{} tasks poses a significant challenge  with sub-problems including visual semantic navigation, object detection, referring expression grounding, and action grounding.
These challenges may be approachable by models that exploit hierarchy~\cite{Bisk:2016ac,kulkarni2016hierarchical}, modularity~\cite{andreas2016neural,eqa_modular}, and structured reasoning and planning~\cite{asai2018classical}.
We are encouraged by the possibilities and challenges that the \dataset{} benchmark introduces to the community.

\section*{Acknowledgements}
Thanks to our UW colleagues for helpful feedback, to Eli VanderBilt and Eric Kolve for their help with \thor{} and leaderboard setup, and Victor Zhong for early modeling design.
And finally, thanks to Ranjay Krishna for sharing the Mechanical Turk annotation interface.
This research was supported in part by the ARO (ARO-W911NF-16-1-0121), the NSF (IIS1252835, IIS-1562364, NSF-NRI-1637479), and the Allen Institute for Artificial Intelligence.


{\small
\bibliographystyle{ieee_fullname}
\bibliography{main}
}

\newpage
\begin{appendices}
~

\setcounter{table}{0}
\setcounter{figure}{0}
\renewcommand{\thetable}{A\arabic{table}}
\renewcommand{\thefigure}{F\arabic{figure}}

\vspace{-23pt}
\section{Dataset Details}
We give additional information about the generation of expert demonstrations in \thor{}, language directives, the annotation interface used to collect directives, and samples of annotations with their associated demonstrations.

\subsection{Expert Demonstrations}

\begin{figure}[b]
    \centering
    \includegraphics[width=0.8\linewidth]{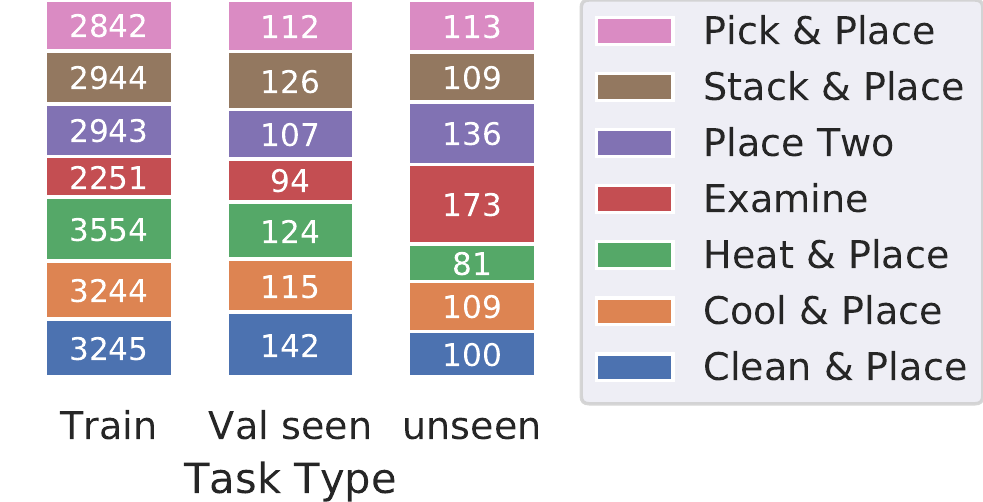}
    \caption{Task distribution across train, validation seen and unseen dataset splits.}
    \label{fig:task_dist}
\end{figure}

\begin{figure}[b]
    \centering
    \includegraphics[width=\linewidth]{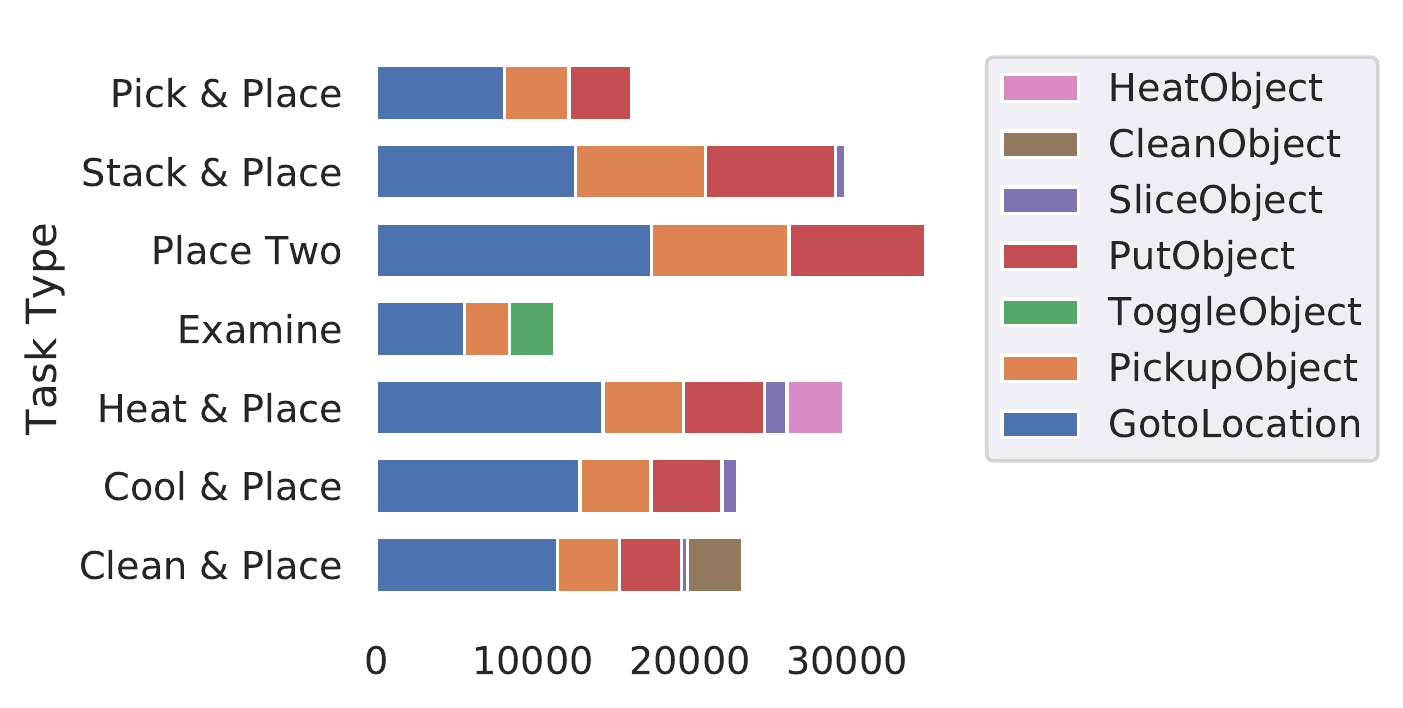}
    \caption{Subgoal distribution across 7 task types.}
    \label{fig:subgoal_dist}
\end{figure}

\begin{figure}[t]
        \includegraphics[width=\linewidth,clip,trim=0 12 0 0]{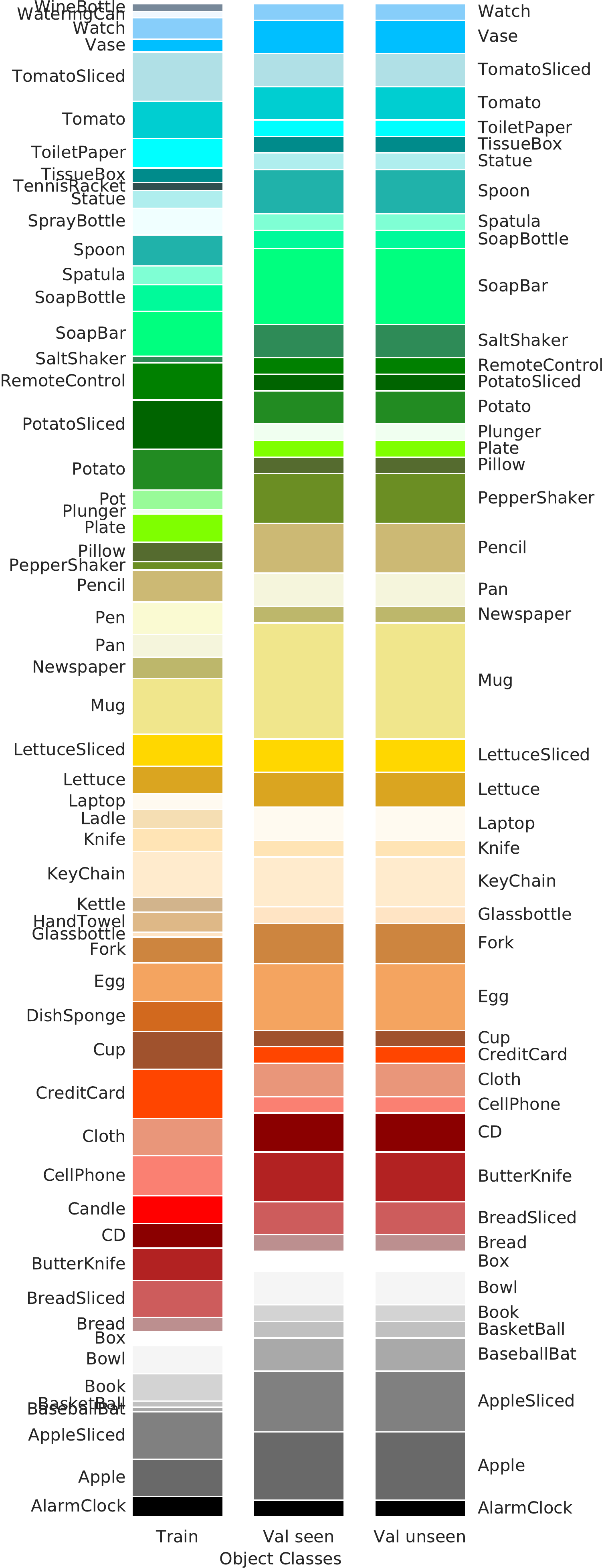}
        \caption{Pickup distributions in the train, validation \textit{seen} and \textit{unseen} folds.}
        \label{fig:objects}
    \end{figure}

\begin{figure}[t]
            \includegraphics[width=\linewidth,clip,trim=0 12 0 0]{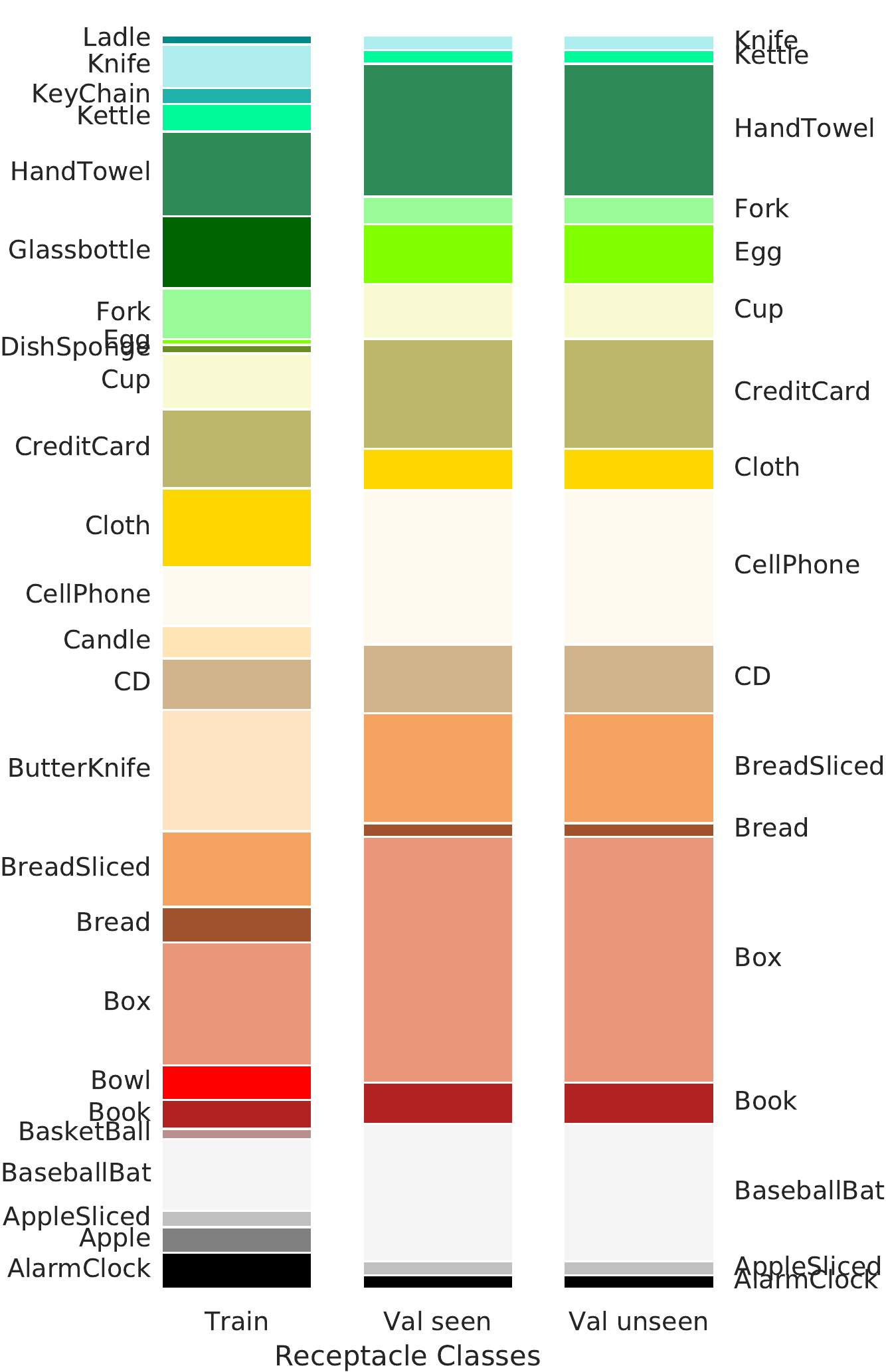}
            \caption{Receptacle distributions in the train, validation \textit{seen} and \textit{unseen} folds.}
            \label{fig:receptacle}
    \end{figure}

When sampling task parameters, we employ an active strategy to maximize data heterogeneity.
Figure~\ref{fig:task_dist} shows the distribution of high-level task across train, validation seen, and validation unseen folds. Figure~\ref{fig:subgoal_dist} shows the distribution of subgoals across task types. And Figures~\ref{fig:objects} and \ref{fig:receptacle} give the distributions of pickup objects and receptacles across the dataset.
Each task parameter sample is defined by $(t, s, o, r, m)$, where 
\begin{itemize}[noitemsep]
    \item $t = $ the task type;
    \item $s = $ the scene in \thor{};
    \item $o = $ the object class to be picked up;
    \item $r = $ the final destination for $o$ or $\emptyset$ for \textbf{Examine};
    \item $m = $ the secondary object class for \textbf{Stack \& Place} tasks ($\emptyset$ for other task types).
\end{itemize}

To construct the next tuple, we first find the largest source of imbalance in the current set of tuples. For example if $o = apple$ is more common than $o = plunger$, $o = plunger$ will be ranked higher than $o = apple$. We additionally account for the prior distribution of each entity (\eg, if \textit{cup} is already represented in the data often as both $o$ and $m$, it becomes disfavored by the sampling algorithm for all slots). We do this greedily across all slots until the tuple is complete. Given any partial piece of information about the task, the distributions of the remaining task parameters remain heterogeneous under this sampling, weakening baseline priors such as ignoring the language input and always executing a common task in the environment. 

Once a task parameter sample is complete, the chosen scene is instantiated, objects and agent start position are randomized, and the relevant room data is encoded into PDDL rules for an expert demonstration.
If the PDDL planner cannot generate an expert demonstration given the room configuration, or if the agent fails an action during execution, for example by running into walls or opening doors onto itself due to physical constraints, the episode is abandoned.
We gather three distinct expert demonstrations per task parameter sample.
These demonstrations are further vetted by rolling them forward using our wrapper to the \thor{} API to ensure that a ``perfect'' model can reproduce the demonstration. The full sampling generation and verification code will be published along with the dataset.

\subsection{Example Language Directives}

\begin{figure}
    \centering
    \includegraphics[width=\linewidth]{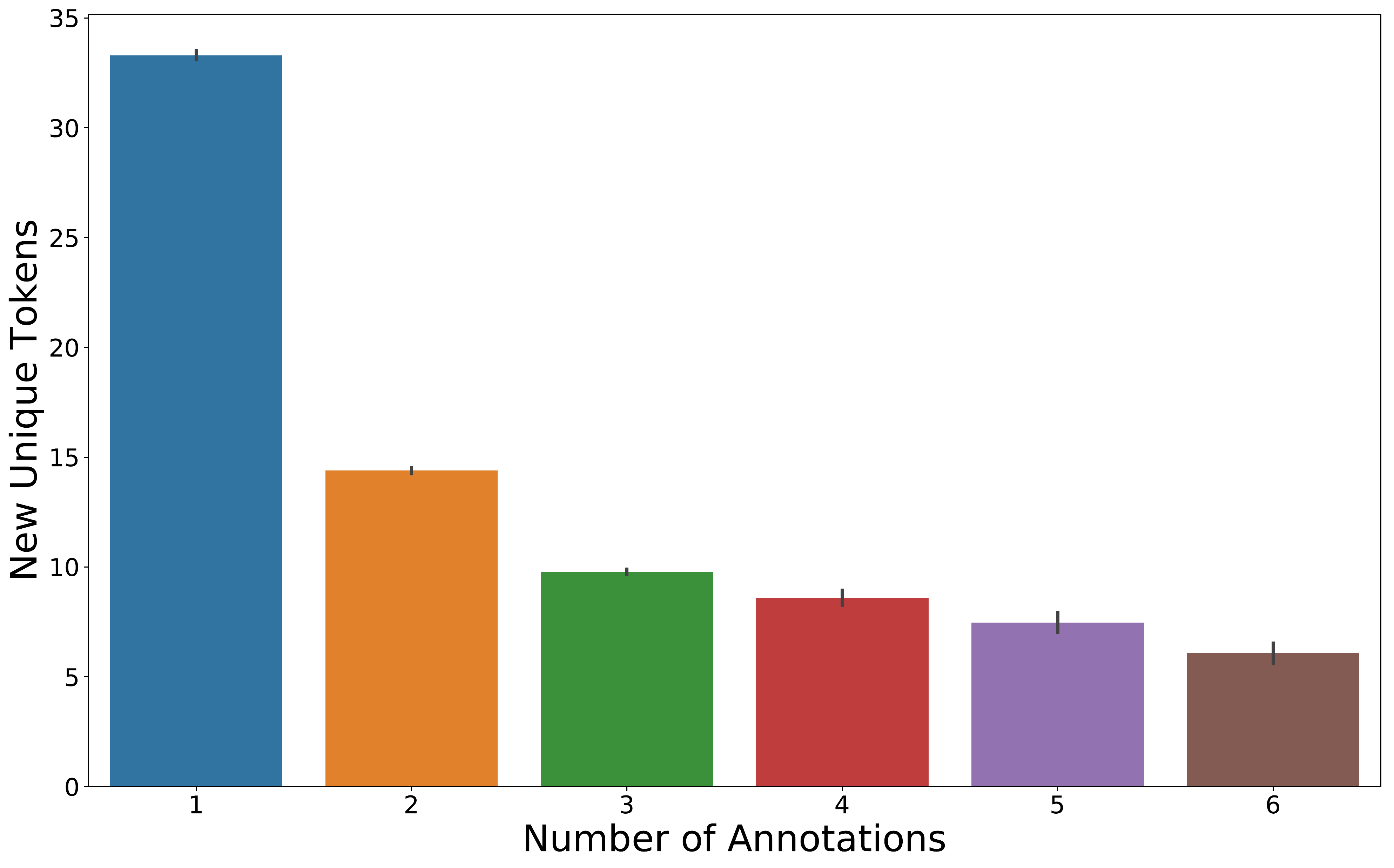}
    \caption{The number of unique tokens introduced per annotation of language directives.}
    \label{fig:annots_vs_unique}
\end{figure}

We chose to gather three directives per demonstration empirically.
For a subset of over 700 demonstrations, we gathered up to 6 language directives from different annotators.
We find that after three annotations, fewer than 10 unique tokens on average are introduced by additional annotators (Figure~\ref{fig:annots_vs_unique}).

\subsection{Annotation Interface}

\begin{figure}
    \centering
    \includegraphics[width=\linewidth]{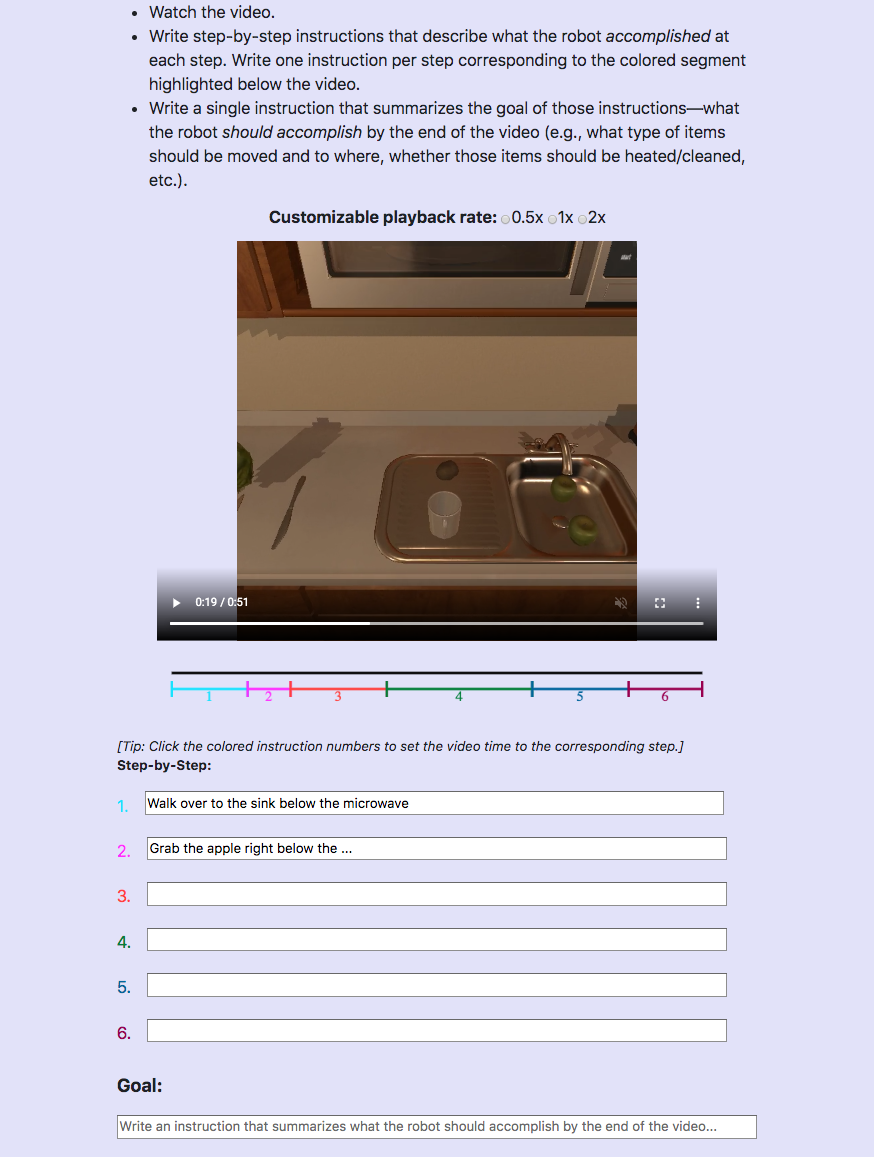}
    \caption{Mechanical Turk Annotation Interface.}
    \label{fig:annotation_interface}
\end{figure}

\figref{fig:annotation_interface} shows the Mechanical Turk interface used to gather language annotations. Workers were presented with a video of the expert demonstration with timeline segments indicating sub-goals. The workers annotated each segment while scrubbing through the video, and wrote a short summary description for the entire sequence. We payed workers \$0.7 per annotation. During vetting, annotators were paid \$0.35 per HIT (Human Interaction Task) to compare 5 sets of three directives each. These wages were set based on local minimum-wage rates and average completion time.

\subsection{Vocabulary Distributions}

\begin{figure*}[t]
    \centering
    \includegraphics[width=\linewidth]{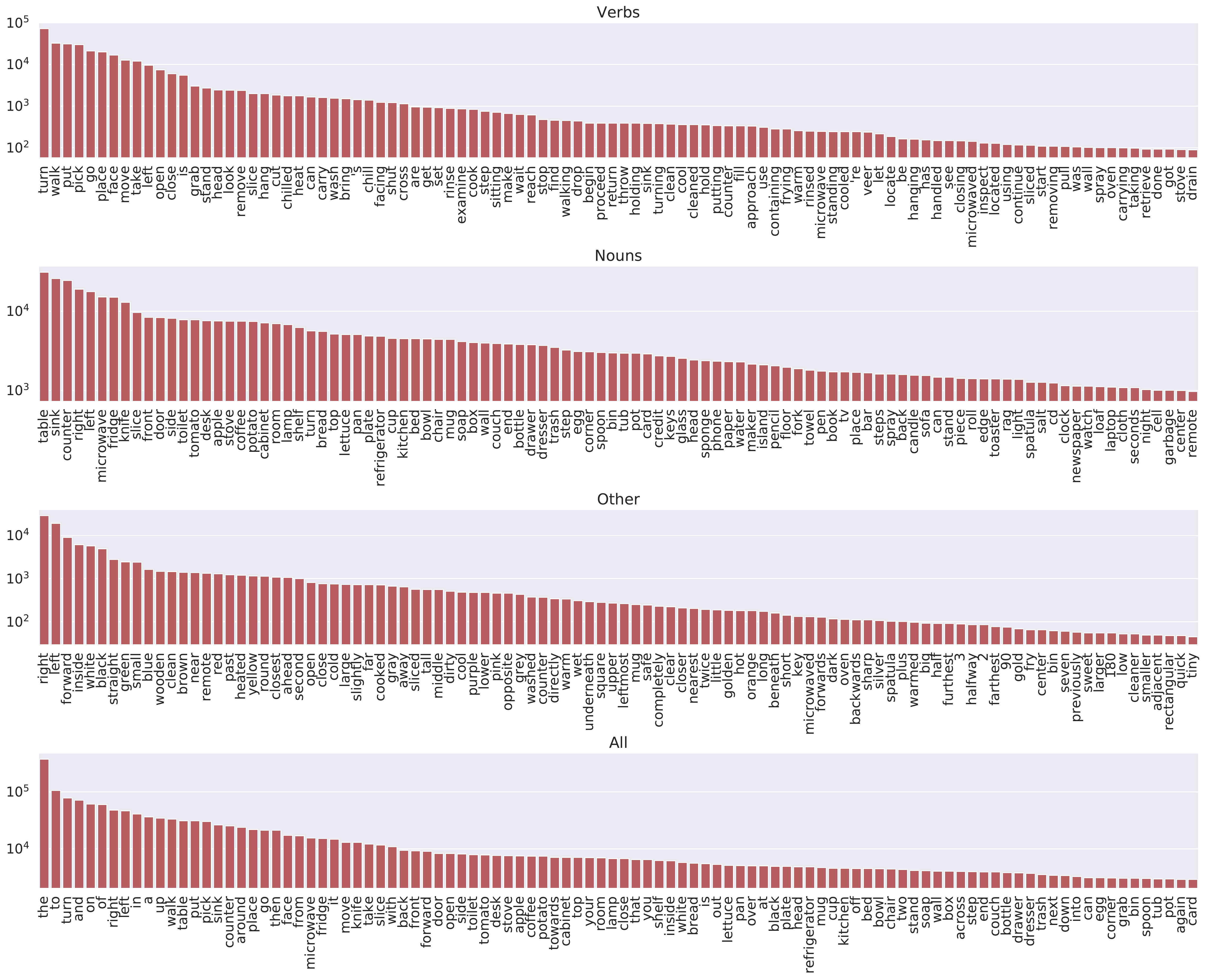}
    \caption{\textbf{Vocabulary Distributions.} Frequency distributions of 100 most common verbs, nouns, other words (non-verbs and non-nouns), and all words.}
    \label{fig:vocab_dist}
\end{figure*}

\figref{fig:vocab_dist} shows vocabulary statistics of the language in~\dataset{}.

\subsection{Dataset Examples}

\figref{fig:long_dataset_examples_pt1} shows 7 expert trajectories (one per task type) and their accompanied annotations.

\section{Implementation Details} \label{sec:implementation_details}

We describe implementation and training details of our baseline Sequence-to-Sequence models.

\paragraph{Preprocessing}
We tokenize the language directives and convert all tokens to lower-case. During dataset generation, we save images from \thor{} $300\times300$ pixels, and later resize them to $224\times224$ during training. The generation pipeline saves initialization information for objects and the agent, so all demonstration can be perfectly replayed in the simulator. Researchers can use this replay feature to augment the dataset by saving high-res images, depth maps, or object-segmentation masks.

\paragraph{Network Architecture}

We use a pretrained ResNet-18 \cite{He_2016_CVPR} to extract $512\times7\times7$ features from the $\texttt{conv5}$ layer. These features are fed into a two-layer CNN with $1\times1$ convolutions to reduce the channel dimension from $512$ to $64$. The $64\times7\times7$ output is flattened, and a fully-connected layer produces a $2500$-dimensional visual feature $v_{t}$.

The language encoder is a bi-directional LSTM with a hidden-dimension of $100$. We do not use pretrained language models to initialize the LSTM, and the encodings are learned from scratch in an end-to-end manner. We also use a self-attention mechanism to attend over the encodings to initialize the hidden-state of the decoder LSTM. 

The action decoder is an LSTM with a hidden-dimension of $512$. The actor is a fully-connected layer that outputs logits for 13 actions. The mask decoder is a three-layer deconvolution network, which takes in the concatenated vector $u_{t}$ and transforms it into $64\times7\times7$ features with a fully-connected layer. These features are subsequently up-scaled into a $1\times300\times300$ binary mask through three layers of deconvolutions and up-sampling with bi-linear interpolation.

\begin{figure}
    \centering
    \includegraphics[width=1.0\linewidth]{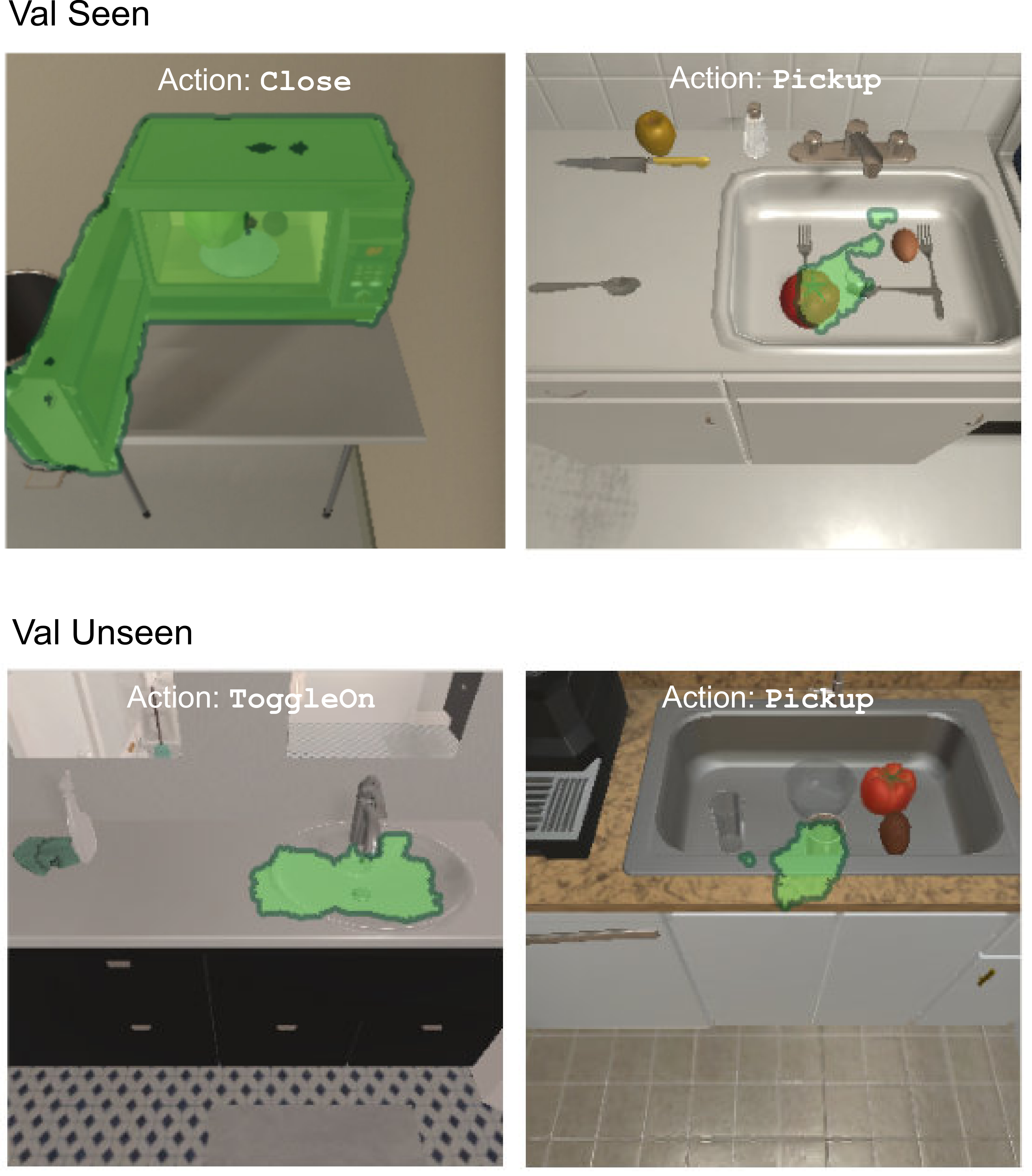}
    \caption{\textbf{Predicted interaction masks.} Masks generated by the \basemodels{Seq2Seq+PM} model are displayed in green.}
    \label{fig:predicted_masks}
\end{figure}

\paragraph{Training}

The models were implemented with PyTorch and trained with the Adam optimizer \cite{kingma2014adam} at a learning rate of 1e-4. We use dropout of $0.3$ on the visual features and the decoder hidden state, tuned on the validation data. Both the action and mask losses are weighted equally, while the auxiliary losses are scaled with a factor of $0.1$. For evaluation, we choose models with the lowest loss on the validation \textit{seen} set. It should be noted that, due to the nature of the tasks, low validation loss might not directly lead to better evaluation performance since the agent does not have to exactly imitate the expert to complete the task.

\paragraph{Notes on Random Agent}
Unlike discretized navigation where taking random actions might allow the agent to stumble upon the goal, \dataset~tasks are much harder to achieve by chance.
The action space branching factor of Room-to-Room navigation~\cite{anderson18}, for example, is $4^6 \approx4000$ ($6$ average steps and $4$ navigation actions).
By contrast, the \dataset{} average branching factor is $12^{50}\approx10^{53}$ ($50$ average steps for $12$ actions).
Beyond action type prediction, the \dataset{} state space resulting from dynamic environments and the need to produce pixel-wise masks for interactive actions explodes further.


\subsection{Predicted Masks}

\figref{fig:predicted_masks} shows a few examples of masks generated by the \basemodels{Seq2Seq+PM} model in seen and unseen validation scenes. The \objects{Microwave} mask accurately captures the contours of the object since the model is familiar with receptacles in seen environments. In contrast, the \objects{Sink} mask in the unseen scene poorly fits the unfamiliar object topology.

\section{Additional Results} \label{additional_res}




\subsection{Performance by Task Type}

In Table~\ref{tab:task_metrics}, we present success rates across the 7 task types. Even the best performing model, \basemodels{Seq2Seq+PM}, mostly succeeds in solving some short-horizon tasks like \tasks{Pick \& Place} and \tasks{Examine}. Long horizon tasks like \tasks{Stack \& Place} and \tasks{Pick Two \& Place} have near zero success rates across all models.

\begin{table}
\centering
\begin{small}
\begin{tabular}{@{}l@{\hspace{6pt}}acacac@{}}
           & \multicolumn{6}{c}{Task Ablations - \textbf{Validation}} \\
           & \mcc{2}{\basemodels{No Language}} & \mcc{2}{\basemodels{Seq2Seq}} & \mcc{2}{\basemodels{Seq2Seq+PM}}    \\
Task Type  & \splits{Seen}   &\splits{Unseen} & \splits{Seen} & \splits{Unseen} & \splits{Seen} & \splits{Unseen} \\
\toprule
Pick \& Place       & 0.0      & 0.0     &    6.3     & \B{1.0}      & \B{7.0}        & 0.0  \\ 
Stack \& Place      & 0.0      & 0.0     &    0.0     & 0.0      & \B{0.9}        & 0.0  \\ 
Pick Two   & 0.0      & 0.0     &    \B{1.6}     & 0.0      & 0.8        & 0.0  \\ 
Clean \& Place      & 0.0      & 0.0     &    0.0     & 0.0      & 1.8        & 0.0   \\  
Heat \& Place       & 0.0      & 0.0     &    \B{1.9}     & 0.0      & \B{1.9}        & 0.0  \\ 
Cool \& Place       & 0.0      & 0.0     &    2.4     & 0.0      & \B{4.0}        & 0.0  \\ 
Examine    & 0.0      & 0.0     &    4.3     & 0.0      & \B{9.6}        & 0.0  \\ 
\bottomrule
\end{tabular}
\end{small}
\caption{\textbf{Success percentages across 7 task types.}
The highest values are shown in \B{blue}.}
\label{tab:task_metrics}
\end{table}


\begin{figure*}[t]
    \centering
    \includegraphics[width=\linewidth]{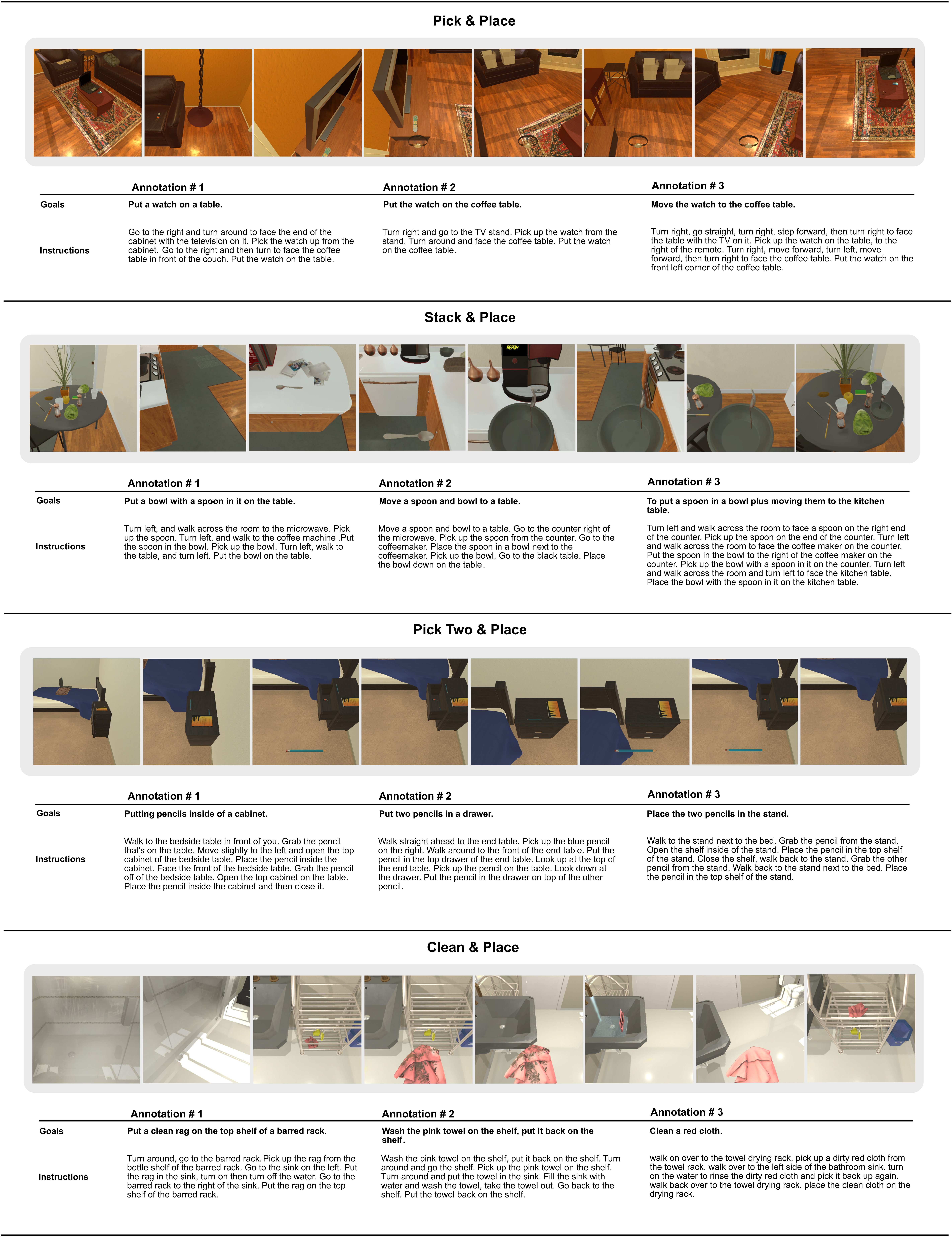}
    \caption{\textbf{Dataset Examples.} Annotations for seven expert demonstrations.}
    \label{fig:long_dataset_examples_pt1}
\end{figure*}

\begin{figure*}[t]\ContinuedFloat
    \centering
    \includegraphics[width=\linewidth]{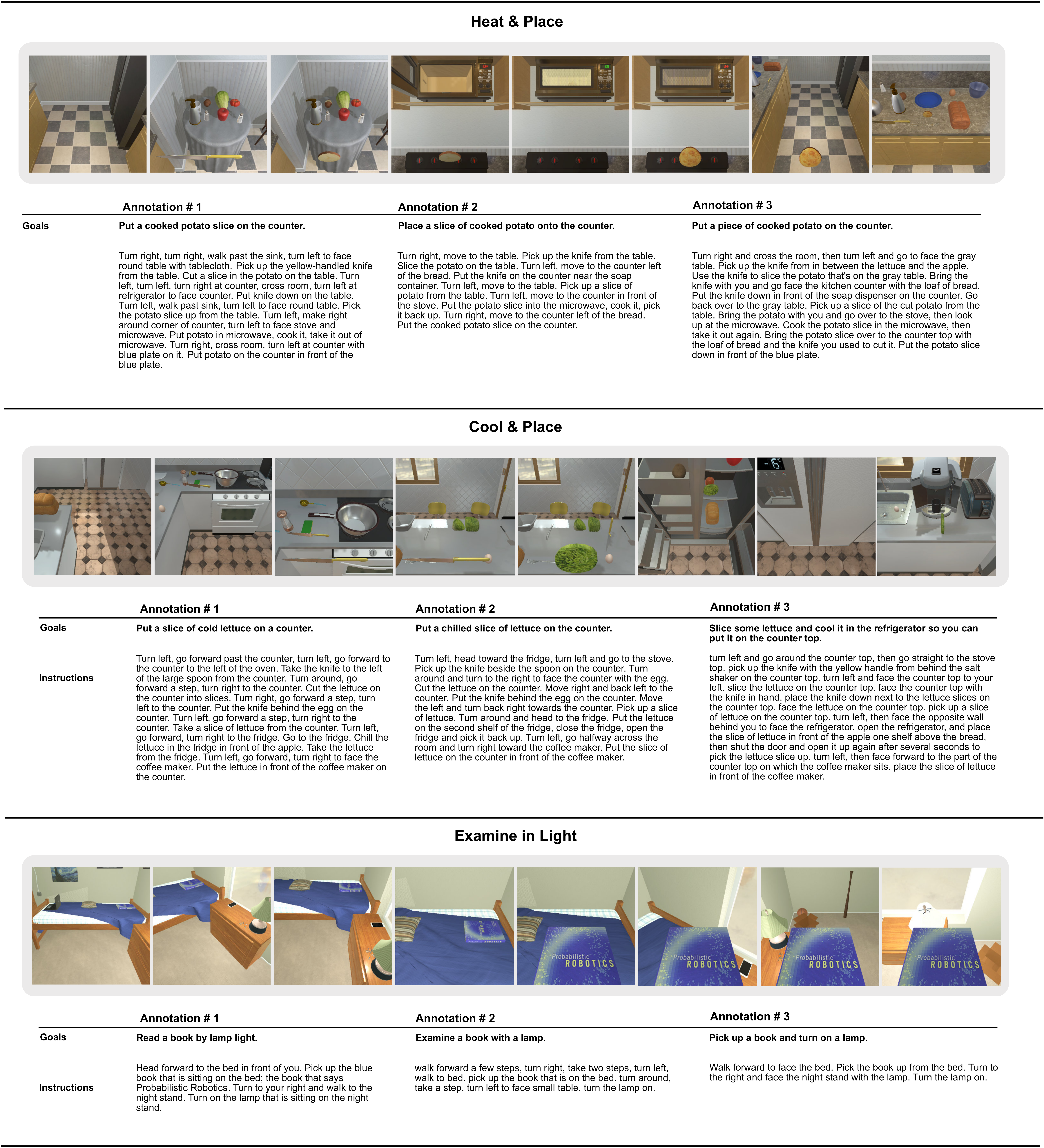}
    \caption{\textbf{Dataset Examples.} Annotations for seven expert demonstrations.}
\end{figure*}

\begin{figure*}[t]
    \centering
    \includegraphics[width=\linewidth]{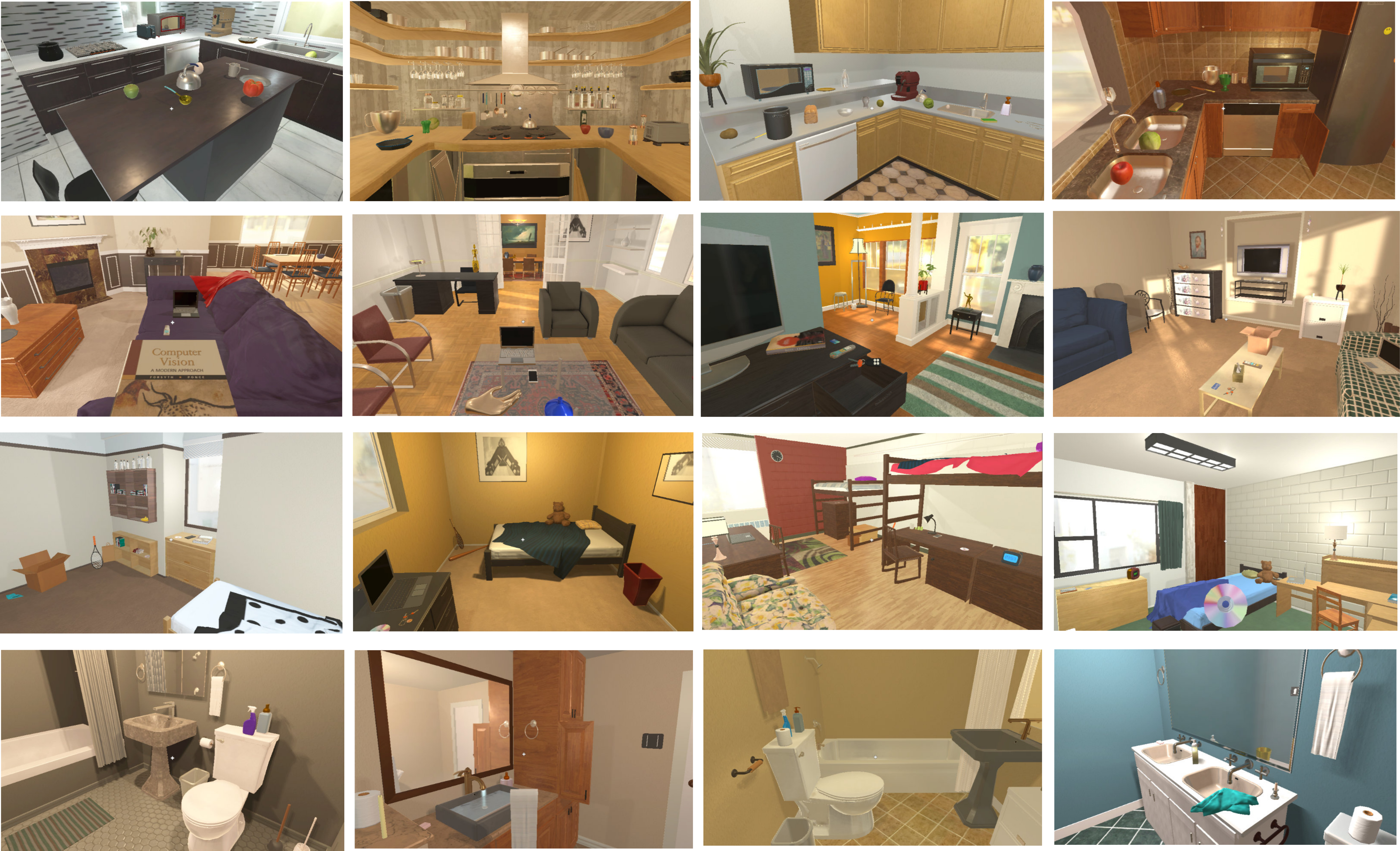}
    \caption{\textbf{Visual diversity of \thor{}~\cite{ai2thor} scenes.} Top to bottom rows: kitchens, living rooms, bedrooms, and bathrooms. Object locations are randomized based on placeable surface areas and class constraints. See \url{https://ai2thor.allenai.org/ithor/demo/} for an interactive demo.}
    \label{fig:thor_env}
\end{figure*}




\end{appendices}

\end{document}